\pdfoutput=1
\documentclass[letterpaper]{article}

\usepackage[preprint]{aaai2027}
\usepackage[hyphens]{url}
\usepackage{graphicx}
\usepackage{natbib}
\usepackage{caption}
\usepackage{subcaption}
\graphicspath{{figures/}}
\usepackage{amsmath, amssymb}
\usepackage{booktabs}
\usepackage{multirow}
\usepackage{xcolor}
\usepackage{enumitem}
\usepackage{placeins}
\usepackage{tikz}
\usetikzlibrary{positioning, arrows.meta, calc}

\setlist[itemize]{topsep=2pt, partopsep=0pt, itemsep=1pt, parsep=0pt, leftmargin=1.3em}
\setlist[enumerate]{topsep=2pt, partopsep=0pt, itemsep=1pt, parsep=0pt, leftmargin=1.5em}
\setlist[description]{topsep=2pt, partopsep=0pt, itemsep=1pt, parsep=0pt}

\title{Beyond Compression: Diagnosing How Post-Training Changes Mathematical Reasoning}

\author{
    Hongyang Li\textsuperscript{\rm 1},
    Yiming Zhu\textsuperscript{\rm 2},
    Xiao Li\textsuperscript{\rm 2},\\
    Caesar Wu\textsuperscript{\rm 1},
    Said Mammar\textsuperscript{\rm 3},
    Pascal Bouvry\textsuperscript{\rm 1}
}
\affiliations{}

\begin{document}
\maketitle

{\renewcommand{\thefootnote}{}%
\footnotetext{\raggedright
\textsuperscript{\rm 1}University of Luxembourg\\
\textsuperscript{\rm 2}Seafill Open-Source Community\\
\textsuperscript{\rm 3}Universit\'e Paris-Saclay\\[2pt]
\{hongyang.li, caesar.wu, pascal.bouvry\}@uni.lu\\
xiao.li@seafill.com \textbullet\ z13655249157@gmail.com\\
said.mammar@univ-evry.fr}}

\begin{abstract}

Post-training is central to mathematical reasoning in modern large language models (LLMs), but endpoint pass@$1$ alone underidentifies what has changed. Gains may reflect newly reachable solutions, cheaper sampling of latent solutions, surface robustness, or memorisation. We compare three post-training paths under a common diagnostic readout: our sufficiently trained off-policy distillation trajectories, released Qwen3 off-policy-plus-on-policy distillation endpoints, and a released DeepSeek-Math endpoint trained with Group Relative Policy Optimisation (GRPO). Our probe uses cross-surface pass@$K$ over verbatim prompts, paraphrases, numerical isomorphisms, and translations, plus consistency, distribution-shape, and verified supervised-fine-tuning (SFT) membership analyses. We find two regimes. On easier AMC problems, large-$K$ ceilings are near saturation, so post-training mainly compresses sample cost. On harder AIME problems, post-training expands the large-$K$ ceiling over the base model: sufficient off-policy distillation already raises this ceiling, Qwen3 released endpoints raise it further, and DeepSeek-Math GRPO does not dominate sufficient off-policy distillation at large $K$. English-dominant distillation improves non-English reasoning but preserves language-tier gaps. A controlled-overfit audit finds limited sensitivity in current SFT-membership probes. Compression is one regime of post-training, not a universal explanation.
\end{abstract}

\section{Introduction}

Open-weight reasoning models now achieve strong scores on mathematical benchmarks, but endpoint pass@$1$ alone does not identify what post-training has changed. The same gain may reflect newly reachable solutions, cheaper sampling of latent solutions, robustness to surface variation, or memorisation of benchmark content. These explanations imply different conclusions for evaluation and deployment, yet they can produce the same endpoint accuracy.

This ambiguity is central to current debates on mathematical post-training. Prior work argues that reinforcement learning (RL) can mainly compress sample cost, raising pass@$1$ without expanding the large-$K$ ceiling~\citep{yue2025does}; other work shows that RL can acquire new compositional behaviours in controlled settings~\citep{yuan2025composition}. Meanwhile, SFT and distillation have become major practical routes to strong reasoning models~\citep{muennighoff2025s1,ye2025limo,guha2025openthoughts}, building on broader knowledge-distillation and reasoning-distillation recipes~\citep{hinton2015distilling,magister2023teaching}; related work further shows that distinct training stages---continual pre-training, problem-solving SFT, and RL---can shape mathematical reasoning differently~\citep{chen2025advancing}. However, static teacher-trace imitation, student-rollout-based teacher matching, and verifier-driven RL need not shape the output distribution in the same way. Surface-robustness studies further show that high reasoning scores can be sensitive to linguistic, symbolic, or functional perturbations~\citep{mirzadeh2024gsm,srivastava2024functional}. Separately, memorisation and membership-inference work shows that language models can retain or reveal training data, but membership is often hard to verify for released models~\citep{carlini2021extracting,carlini2023quantifying}. Existing evaluations therefore often isolate one axis at a time---mechanism, surface form, endpoint, or membership assumption---making these explanations difficult to separate under a common protocol.

\begin{figure*}[!t]
\centering
\includegraphics[width=0.92\textwidth,height=0.20\textheight,keepaspectratio]{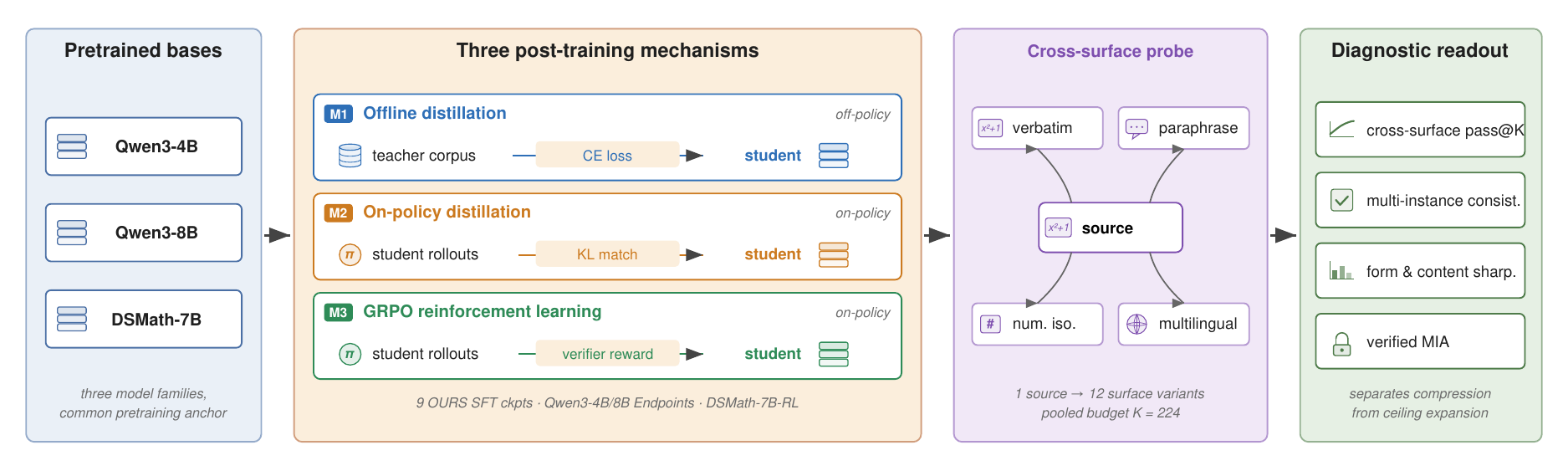}
\caption{\textbf{Overview.} We compare three post-training
paths---our off-policy distillation trajectories (M1), released Qwen3
off-policy-plus-on-policy distillation endpoints (M2), and the released
DeepSeek-Math GRPO endpoint (M3)---on shared base model families.
We evaluate them with cross-surface pass@$K$ over verbatim prompts,
paraphrases, numerical isomorphisms, and translations, together with
strategy-diversity, multi-instance consistency, and verified
memorisation probes.}
\label{fig:overview}
\end{figure*}

We address this with a trajectory-aware diagnostic. The key idea is to evaluate performance as both the sampling budget and problem surface vary. If a gain appears at small $K$ but vanishes at large $K$, it indicates sample-cost compression. If it persists at large $K$, it indicates ceiling expansion: the post-trained model reaches correct solutions that the base model did not reach under the evaluated budget. If the gain fails under paraphrase, numerical isomorphism, or translation, it is surface-fragile. If verified training items separate from verified unseen controls, memorisation becomes a plausible explanation. We compare three post-training paths under this common readout (Fig.~\ref{fig:overview}). First, we train reproducible off-policy distillation trajectories for Qwen3-4B, Qwen3-8B, and DeepSeek-Math-7B. We treat these sufficiently trained off-policy endpoints as first-class post-training models rather than weak SFT baselines. Second, we evaluate released Qwen3-4B and Qwen3-8B endpoints, which use off-policy followed by on-policy distillation for these sizes~\citep{qwen3}. Third, we evaluate the released DeepSeek-Math-7B-RL endpoint as a GRPO comparison~\citep{deepseekmath}. Since some endpoints are released rather than retrained by us, our goal is mechanism-relative diagnosis under a shared protocol, not a perfectly matched causal ablation of every algorithm.

Our main probe is cross-surface pass@$K$. Starting from a shared $100$-problem pool, we pool attempts at the source-problem level, so large-$K$ performance measures whether the model can solve the underlying problem across surface changes, not merely resample one familiar prompt. We complement this with multi-instance consistency, distribution-shape descriptors separating form sharpening from answer-content sharpening, cross-lingual analysis, and verified SFT-membership probes enabled by our controlled training data.

Our results show two regimes. On easier AMC problems, large-$K$ ceilings are near saturation, so post-training mainly behaves as sample-cost compression: correct solutions become cheaper to sample, while mechanisms converge near the ceiling. On harder AIME problems, however, post-training expands the ceiling over the base model. Sufficient off-policy distillation already raises the cross-surface large-$K$ ceiling; released Qwen3 off-policy-plus-on-policy endpoints raise it further; and the released DeepSeek-Math GRPO endpoint does not exceed our sufficiently trained off-policy endpoint at large $K$. Thus, compression explains one regime of post-training, but not all mechanisms or difficulties.

Robustness and distributional diagnostics refine this picture. Qwen3 endpoints strongly consolidate solved problems across paraphrases and numerical isomorphisms, while the DeepSeek-Math GRPO endpoint remains close to its SFT checkpoints in multi-instance consistency. Distribution-shape metrics show that form sharpening and answer-content sharpening can diverge. Cross-lingually, English-dominant distillation improves every non-English language we evaluate, but preserves the stronger-versus-weaker language gap. Finally, a controlled-overfit memorisation audit finds limited sensitivity in current behavioural and forward-pass membership probes for SFT-specific membership.
Our contributions are:
\begin{enumerate}[leftmargin=0.45cm, nosep]
\item \textbf{Mechanism-relative post-training comparison.}
We place sufficiently trained off-policy distillation, released on-policy distillation, and released GRPO endpoints under a common diagnostic readout, while distinguishing controlled trajectories from observed endpoints.
\item \textbf{Cross-surface pass@$K$.}
We introduce a trajectory-aware probe that separates sample-cost compression, large-$K$ ceiling expansion, and surface robustness.
\item \textbf{Regime-dependent view of compression.}
We show that compression explains saturated easy problems, while hard problems reveal ceiling expansion from post-training, already under sufficient off-policy distillation.
\item \textbf{Distributional and robustness diagnostics.}
We connect capability changes to multi-instance consistency, form/content sharpening, multilingual surface shifts, and verified SFT-membership controls.
\end{enumerate}

\section{Method}
\label{sec:method}
The paper is organised around three post-training mechanisms observed in modern open-weight LLMs for mathematical reasoning (\S\ref{sec:method:mechanisms}). We train a sequence of checkpoints under one of them---off-policy distillation---across three
model families (\S\ref{sec:method:sft}). All checkpoints are evaluated with a capability probe set (\S\ref{sec:method:probes}). Sampling, scoring, and strategy-diversity protocols are summarised in \S\ref{sec:method:sampling}.

\subsection{Three post-training mechanisms} 
\label{sec:method:mechanisms} Recent open-weight math reasoning models build on pretrained bases through different post-training mechanisms. We group them into three types, which differ in the training data, objective, and whether updates depend on the student's own rollouts. \begin{description}[ leftmargin=0.55cm, topsep=1pt, itemsep=1pt, parsep=0pt, partopsep=0pt ] \item[(M1) Off-policy distillation.] The student is trained on a static dataset of (input, \emph{teacher output}) pairs with cross-entropy loss on teacher tokens; the student's own distribution does not enter the update \citep{hinton2015distilling}. This covers long chain-of-thought (CoT)~\citep{wei2022chain} SFT recipes used in academic distillation studies~\citep{magister2023teaching,ho2023largelm} and open pipelines such as OpenR1~\citep{openr1math}, NuminaMath~\citep{numinamath}, and DeepSeek-R1 distillations~\citep{deepseekr1}. \item[(M2) On-policy distillation.] The student generates rollouts, and teacher logits provide a KL-matching target for the generated tokens. The Qwen3 report (\citealp{qwen3}, \S4.5) states that the smaller dense models (0.6B--14B) and Qwen3-30B-A3B use off-policy distillation followed by on-policy distillation against 32B/235B teachers, rather than the four-stage SFT+RL pipeline used for flagship models. Thus, the Qwen3-4B and Qwen3-8B Official endpoints we evaluate are on-policy-distilled, not RL-trained. \item[(M3) Reinforcement learning.] The student generates rollouts scored by a rule-based verifier, and policy-gradient updates maximise expected verified reward. We use DeepSeek-Math-7B-RL~\citep{deepseekmath} as our M3 endpoint, whose GRPO recipe starts from an SFT'd DeepSeek-Math-7B checkpoint. \end{description} 
All comparisons are anchored at released base checkpoints within the same family, so post-training paths share a common pretraining reference even when endpoint trajectories are observed rather than controlled. Table~\ref{tab:roster} lists all 15 checkpoints with mechanism labels: our 9 M1 checkpoints provide the open-replicable trajectory, while M2 and M3 endpoints are observed endpoints.

\begin{table}[!htb]
\centering
\footnotesize
\setlength{\tabcolsep}{5pt}
\resizebox{\columnwidth}{!}{%
\begin{tabular}{@{}lll@{}}
\toprule
\textbf{Checkpoint} & \textbf{Post-training pipeline} & \textbf{Source} \\
\midrule
Qwen3-4B-Base                & --                              & Qwen \\
Qwen3-4B-SFT-Math-45k-ep1    & Off-policy distill              & Ours \\
Qwen3-4B-SFT-Math-45k-ep2    & Off-policy distill              & Ours \\
Qwen3-4B-SFT-Math-45k-ep3    & Off-policy distill              & Ours \\
Qwen3-4B (Official)          & Off-policy + On-policy distill  & Qwen \\
\midrule
Qwen3-8B-Base                & --                              & Qwen \\
Qwen3-8B-SFT-Math-90k-ep1    & Off-policy distill              & Ours \\
Qwen3-8B-SFT-Math-90k-ep2    & Off-policy distill              & Ours \\
Qwen3-8B-SFT-Math-90k-ep3    & Off-policy distill              & Ours \\
Qwen3-8B (Official)          & Off-policy + On-policy distill  & Qwen \\
\midrule
DeepSeek-Math-7B-Base               & --                       & DeepSeek \\
DeepSeek-Math-7B-SFT-hybrid-ep1     & Off-policy distill       & Ours \\
DeepSeek-Math-7B-SFT-hybrid-ep2     & Off-policy distill       & Ours \\
DeepSeek-Math-7B-SFT-hybrid-ep3     & Off-policy distill       & Ours \\
DeepSeek-Math-7B-RL                 & Off-policy distill + RL  & DeepSeek \\
\bottomrule
\end{tabular}%
}
\caption{Model roster. Nine checkpoints (\textit{Ours}) instantiate
off-policy distillation. In the \textit{Source} column, \textit{Ours} marks
checkpoints we trained; the rest are released endpoints. Where a released
pipeline is listed as off-policy distillation, that stage is the vendor's own,
not one of ours: DeepSeek-Math-7B-RL applies GRPO on top of DeepSeek's SFT
checkpoint, and the Qwen3 Official endpoints distil off-policy then on-policy.}
\label{tab:roster}
\end{table}

\subsection{Our off-policy distillation training}
\label{sec:method:sft}
\begin{figure}[!htb]
\small
\fbox{\parbox{0.93\columnwidth}{%
\textbf{User:} Find all real $x$ such that $x^{4} - 8x^{2} + 15 = 0$.

\medskip
\textbf{Assistant:}\\[0.3em]
\texttt{<think>}\\
Let $y = x^{2}$. Then $y^{2} - 8y + 15 = (y-3)(y-5) = 0$,
so $y \in \{3,5\}$. Therefore $x^{2} \in \{3,5\}$, giving the
four real roots $x \in \{\pm\sqrt{3},\,\pm\sqrt{5}\}$.\\
\texttt{</think>}

\medskip
The four real roots are $\pm\sqrt{3}$ and $\pm\sqrt{5}$. The final
answer is $\boxed{\{-\sqrt{5},\,-\sqrt{3},\,\sqrt{3},\,\sqrt{5}\}}$.
}}
\caption{A representative training sample for off-policy distillation
(stylised; real samples average $\sim$3k tokens of
$\langle$\texttt{think}$\rangle$ content).}
\label{fig:sample}
\end{figure}

\paragraph{Training corpora.}
For the Qwen3 family we use \textbf{OpenR1-Math}~\citep{openr1math}, a public corpus of competition-level math problems paired with reasoning traces distilled from DeepSeek-R1. We retain samples with exactly one $\langle$\texttt{think}$\rangle\ldots\langle$\texttt{/think}$\rangle$ block, a $\backslash\texttt{boxed}\{\cdot\}$ final answer (Figure~\ref{fig:sample}), and total assistant length below $60$k characters (typical samples
$\approx$6k tokens after Qwen3 tokenization; tail up to $\approx$26k). We draw two splits under a fixed seed: OpenR1-Math-45k for the 4B model and OpenR1-Math-90k for the 8B model; the size difference accommodates the larger model's compute budget. For DeepSeek-Math-7B, whose native context window is only $4{,}096$ tokens, we construct a 4k-fitted hybrid corpus by combining two sources, filtered OpenR1-Math samples and short-form CoT traces sourced from NuminaMath-CoT~\citep{numinamath}. 
\paragraph{Training procedure.}
All checkpoints are full-parameter fine-tunes using the \texttt{verl} SFT trainer~\citep{verl} under Fully Sharded Data Parallel (FSDP) with bfloat16 weights and activation checkpointing. Hardware is four nodes of four
NVIDIA A100-40GB ($16$ GPUs) for the 4B and 7B runs and eight nodes ($32$ GPUs) for the 8B runs. Per-GPU micro-batch is fixed at $1$, sequence packing is disabled, and validation runs on a held-out $200$-sample split every $100$ steps. Checkpoints are saved at every epoch boundary; downstream evaluation uses epoch checkpoints to keep the trajectory axis clean. Total wall-clock per 3-epoch run is $\sim$14 h on Qwen3-4B (16 A100), $\sim$20 h on Qwen3-8B (32 A100), and $\sim$8 h on DeepSeek-Math-7B (16 A100, native 4k context). The full hyperparameter list, training/validation loss curves, and learning-rate / gradient-norm trajectories are reported in Appendix~\ref{app:training} (Table~\ref{tab:hyper}; Figures~\ref{fig:loss_curves}, \ref{fig:lr_curves}, \ref{fig:grad_norm}).
\subsection{Capability probe set (T0--T3)}
\label{sec:method:probes}

\begin{table}[!htb]
\small
\resizebox{\columnwidth}{!}{%
\begin{tabular}{@{}lllr@{}}
\toprule
\textbf{Probe} & \textbf{Design} & \textbf{Items} & \textbf{Rollouts} \\
\midrule
T0   & verbatim                       & $100$ & $4{,}800$ \\
T1   & paraphrase ($3\times$)         & $300$ & $4{,}800$ \\
T2   & numerical isomorphism ($3\times$) & $300$ & $4{,}800$ \\
T3   & translation ($5$ langs)        & $500$ & $8{,}000$ \\
\bottomrule
\end{tabular}%
}
\caption{Capability probe inventory. T0--T3 transform a shared
$100$-problem source pool---$30$ problems from AIME 2025,
$30$ from AIME 2026, and $40$ from AMC 2023. Rollout counts are
per evaluated checkpoint.}
\label{tab:capprobes}
\end{table}

Conventional math benchmarks report pass@$K$ on a single verbatim
problem statement, conflating ``can the model solve this surface
form'' with ``can the model solve this class of problem''---a gap
that surface-form memorisation or distillation leakage can hide. We
therefore instantiate a verbatim anchor plus three perturbation axes
around each source problem and ask whether capability survives each:

\begin{description}[leftmargin=0.55cm, topsep=2pt, itemsep=1pt, parsep=2pt]
\item[T0 (verbatim).] Original English---the conventional benchmark.
\item[T1 (paraphrase).] $3$ length-controlled English paraphrases per
source preserving numerical content and ground-truth answer. Probes
surface-form robustness.
\item[T2 (numerical isomorphism).] $3$ digit-preserving constant
permutations per source; reasoning structure preserved but the
ground-truth answer changes, ruling out answer-level memorisation.
\item[T3 (cross-lingual).] $5$ translations covering Chinese,
Spanish, Swahili, Urdu, and Arabic. Probes cross-lingual
generalisation.
\end{description}

The full probe set contains $1{,}200$ variants per model
($100\!\cdot\!(1+3+3+5)$): T0 is sampled $K\!=\!48$ times per source
(single variant); T1/T2/T3 are sampled $K\!=\!16$ per variant,
giving source-level pools of $48$ on T1/T2, $80$ on T3, and combined
pass@$K$ budgets of $96$ (T1$\cup$T2), $144$ (T0$\cup$T1$\cup$T2),
and $224$ (full T0$\cup$T1$\cup$T2$\cup$T3). An honest capability
gain should propagate beyond the verbatim anchor; surface-fragile
gains register preferentially on T0. The inventory is shown in
Table~\ref{tab:capprobes}.

\subsection{Sampling and strategy-diversity}
\label{sec:method:sampling}

All rollouts use vLLM~\citep{kwon2023efficient} with $T = 0.7$, top-$p = 0.95$, and
$\texttt{max\_new\_tokens} = 16{,}384$ (Qwen3) or $4{,}096$ (DSMath
native context). Responses are scored by a balanced-brace
$\backslash\texttt{boxed}\{\cdot\}$ parser with LaTeX equivalence
(Appendix Fig.~\ref{fig:qa_example} shows a representative rollout
with full reasoning trace). We also test whether each post-training mechanism sharpens
$p(y\mid x)$~\citep{yue2025does,kirk2024rlhf} at three layers:
answer-level (self-consistency majority share~\citep{wang2023self},
distinct-answer count, entropy), response-level (pairwise $4$-gram
Jaccard~\citep{li2016diversity}, MiniLM-L6~\citep{wang2020minilm} embedding cosine
distance, length CV), and cluster-level (single-link clustering at
cosine $\varepsilon\!=\!0.3$). The layers separate form-collapse
(templates) from content-collapse (answers)~\citep{stanton2021distillation}.
\section{Cross-mechanism analysis}
\label{sec:sharpening}
We evaluate all three on two orthogonal axes: \emph{capability} (the highest pass@$K$ a model reaches when given $K$ surface-perturbed attempts at a problem) and \emph{mechanism signature} (the shape of the output distribution that produces those attempts).

\paragraph{Overview.}
\label{sec:mech:overview}
For each source problem the probe set yields an attempt budget,
pooled at the source-problem level, of
\[
K_{\max} = \underbrace{1{\times}48}_{\text{T0}}
         + \underbrace{3{\times}16}_{\text{T1}}
         + \underbrace{3{\times}16}_{\text{T2}}
         + \underbrace{5{\times}16}_{\text{T3}} = 224,
\]
where T0 is the original problem and T1, T2, and T3 are variants. The
cross-surface pass@$K$ at $K{=}224$ is therefore the probability
that \emph{any} of $224$ attempts---spanning paraphrase, numerical
isomorphism, and translation---produces a correct
answer for that source problem. This probe extends the
capability-ceiling pass@$K$ of \citet{yue2025does} from i.i.d.\
resampling on a single prompt to $12$ surface variants per source
problem, which is a \emph{strictly stronger} capability
probe---i.i.d.\ resampling on the original prompt is the T0 special
case ($K\leq 48$) of our $K=224$ budget, and a model must solve
the underlying problem under \emph{any} surface to score
(formal definition in Appendix~\ref{app:formal_passk}).
\subsection{Compression vs.\ ceiling expansion}
\label{sec:mech:ceiling}
\begin{figure}[!htb]
\centering
\begin{subfigure}[t]{0.49\linewidth}
  \centering
  \includegraphics[width=\linewidth]{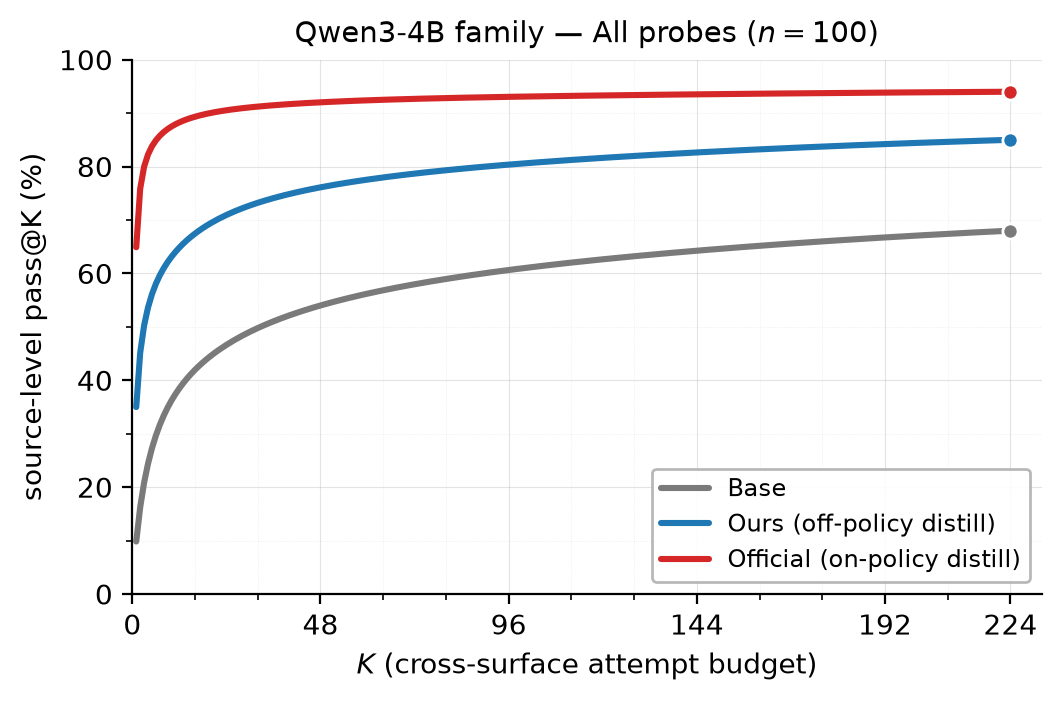}
  \caption{4B, full set.}
  \label{fig:passK_qwen3:all_4b}
\end{subfigure}\hfill
\begin{subfigure}[t]{0.49\linewidth}
  \centering
  \includegraphics[width=\linewidth]{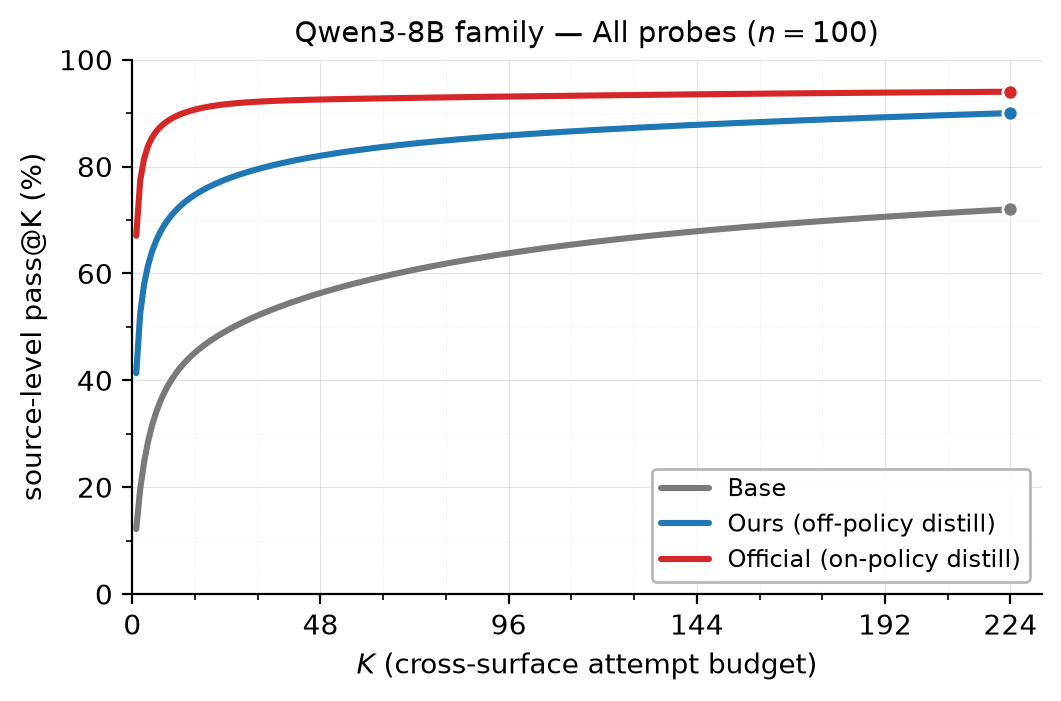}
  \caption{8B, full set.}
  \label{fig:passK_qwen3:all_8b}
\end{subfigure}
\\[2pt]
\begin{subfigure}[t]{0.49\linewidth}
  \centering
  \includegraphics[width=\linewidth]{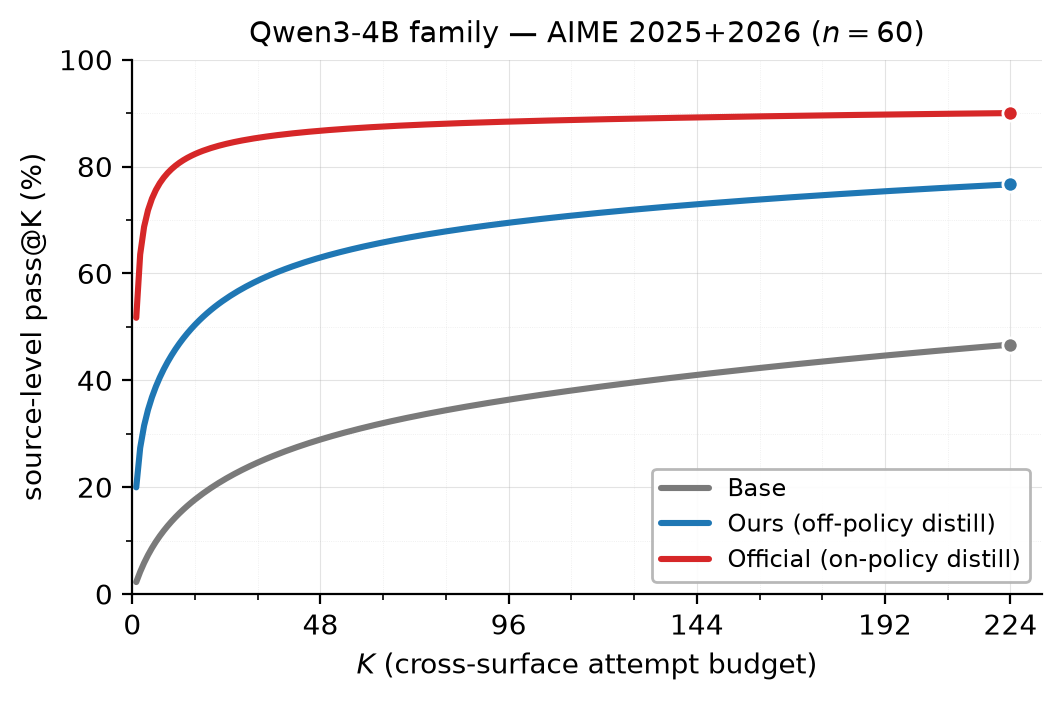}
  \caption{4B, AIME ($n{=}60$).}
  \label{fig:passK_qwen3:aime_4b}
\end{subfigure}\hfill
\begin{subfigure}[t]{0.49\linewidth}
  \centering
  \includegraphics[width=\linewidth]{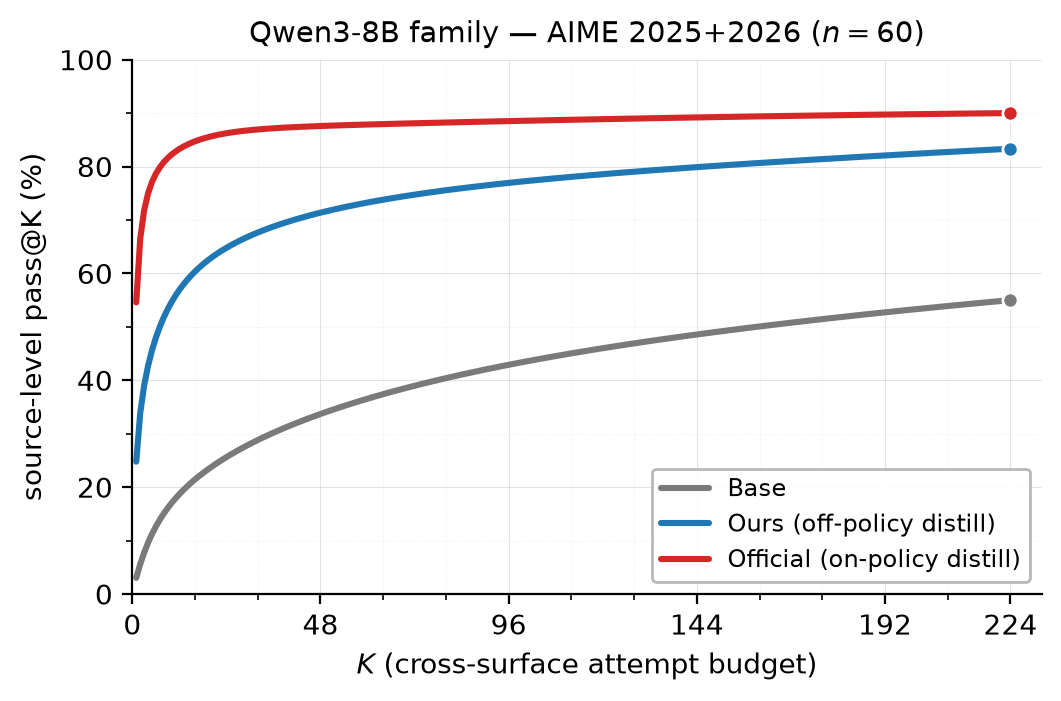}
  \caption{8B, AIME ($n{=}60$).}
  \label{fig:passK_qwen3:aime_8b}
\end{subfigure}
\\[2pt]
\begin{subfigure}[t]{0.49\linewidth}
  \centering
  \includegraphics[width=\linewidth]{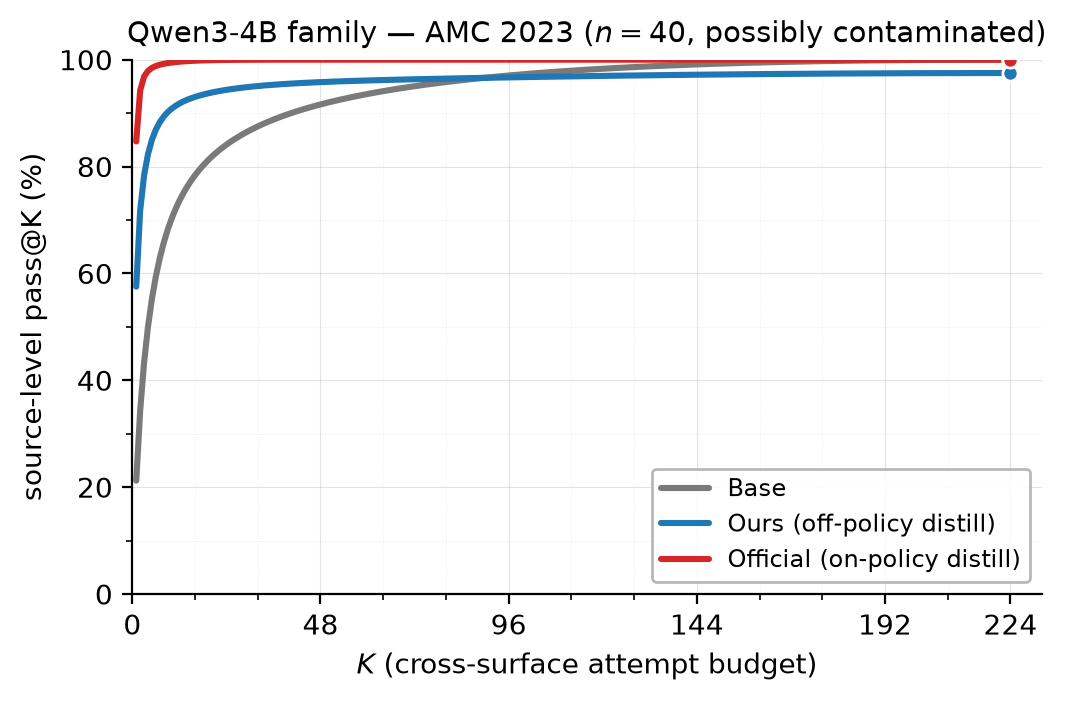}
  \caption{4B, AMC ($n{=}40$).}
  \label{fig:passK_qwen3:amc_4b}
\end{subfigure}\hfill
\begin{subfigure}[t]{0.49\linewidth}
  \centering
  \includegraphics[width=\linewidth]{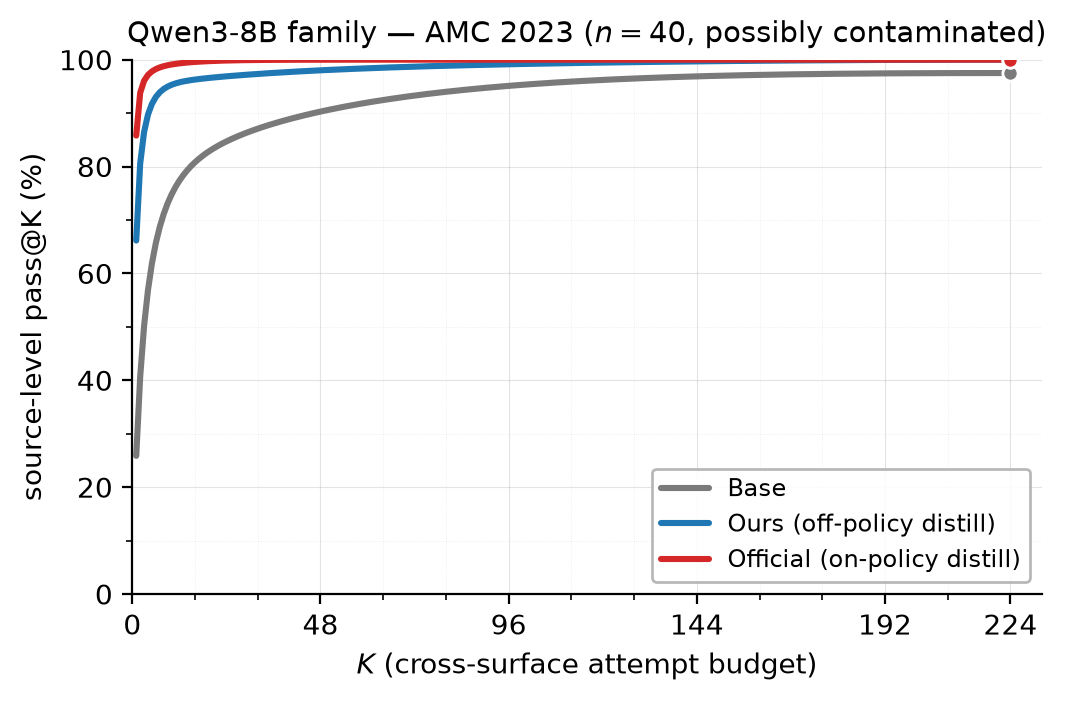}
  \caption{8B, AMC ($n{=}40$).}
  \label{fig:passK_qwen3:amc_8b}
\end{subfigure}
\caption{\textbf{Qwen3 cross-surface pass@$K$.}}
\label{fig:passK_qwen3}
\end{figure}

\begin{figure}[!htb]
\centering
\begin{subfigure}[t]{0.30\linewidth}
  \centering
  \includegraphics[width=\linewidth]{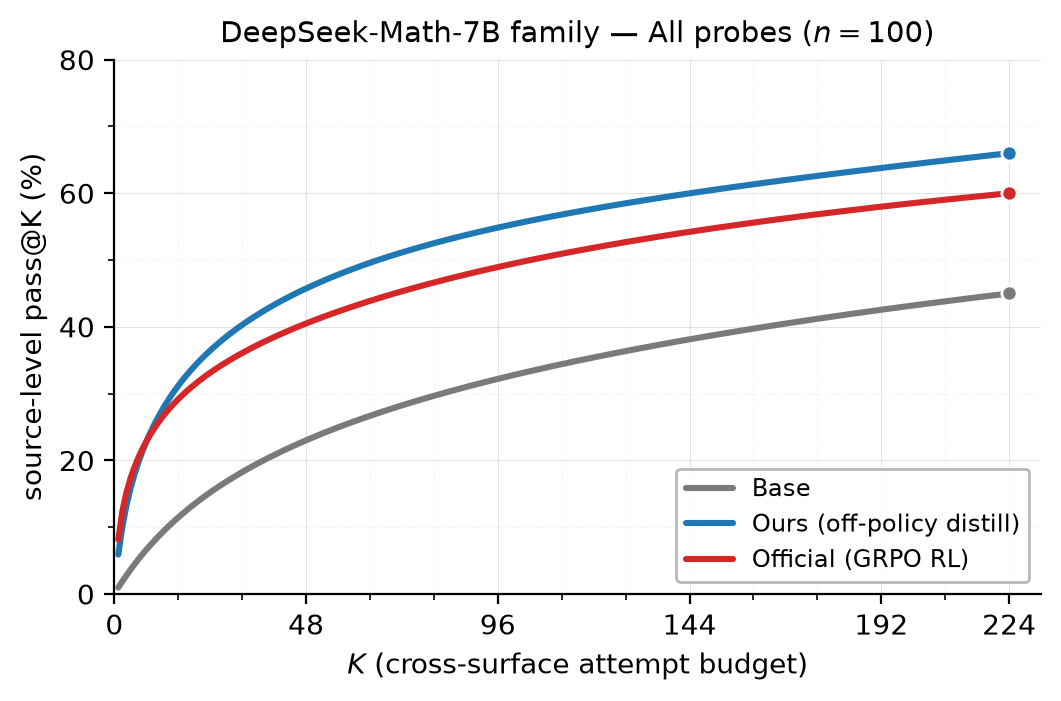}
  \caption{Full $n{=}100$.}
  \label{fig:passK_dsmath:all}
\end{subfigure}\hfill
\begin{subfigure}[t]{0.34\linewidth}
  \centering
  \includegraphics[width=\linewidth]{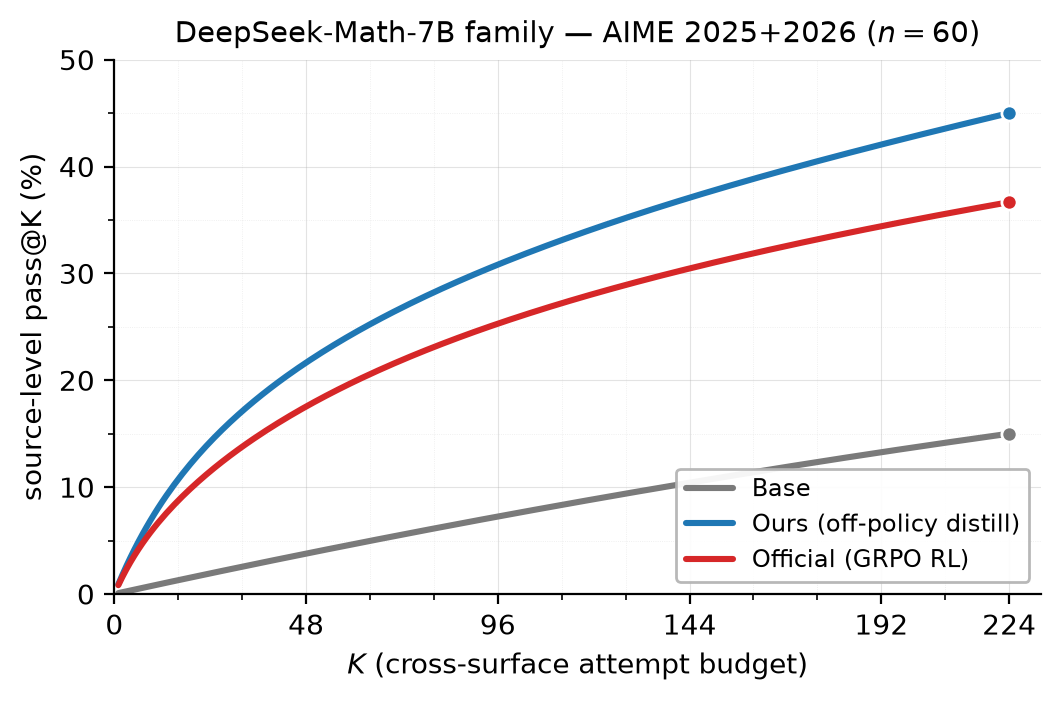}
  \caption{AIME ($n{=}60$).}
  \label{fig:passK_dsmath:aime}
\end{subfigure}\hfill
\begin{subfigure}[t]{0.32\linewidth}
  \centering
  \includegraphics[width=\linewidth]{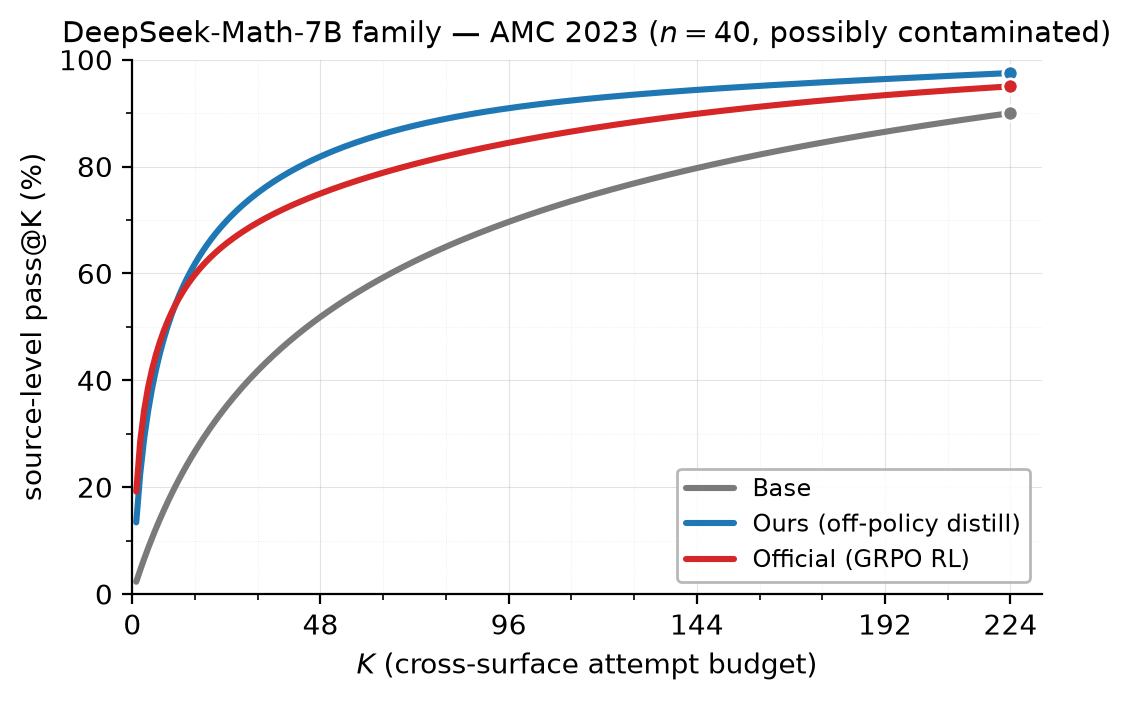}
  \caption{AMC ($n{=}40$).}
  \label{fig:passK_dsmath:amc}
\end{subfigure}
\caption{\textbf{DeepSeek-Math-7B cross-surface pass@$K$.}
Base, our off-policy distillation checkpoint, and the released GRPO
endpoint are evaluated under the same pooled T0--T3 budget.}
\label{fig:passK_dsmath}
\end{figure}

The full cross-mechanism capability summary (pass@1, pass@16, pass@48,
pass@96, pass@144, pass@224 for all 15 checkpoints) is reported in
Table~\ref{tab:capability} (Appendix~\ref{app:subset_tables}). We
read it along two axes: \emph{difficulty}
(AMC 2023 vs.\ the harder AIME 2025+2026 subset) and \emph{mechanism}
(off-policy distillation alone vs.\ off-policy + on-policy distillation
vs.\ off-policy + GRPO RL). Figures~\ref{fig:passK_qwen3} and
\ref{fig:passK_dsmath} plot the full $pass@K$ trajectories;
per-subset numeric anchor points at
$K\in\{1, 48, 96, 144, 224\}$ for all three families are tabulated
in Appendix~\ref{app:subset_tables}
(Tables~\ref{tab:qwen3_aime}--\ref{tab:dsmath_amc}).

\paragraph{Easy regime (AMC): mechanism differences collapse at the ceiling.} 
On AMC 2023, $pass@224$ is already near saturation across families: $100/98/100$ for Qwen3-4B, $98/100/100$ for Qwen3-8B, and $90/98/95$ for DSMath-7B under Base/SFT-ep3/Endpoint (Tables~\ref{tab:qwen3_amc}, \ref{tab:dsmath_amc}). Where headroom remains, SFT closes most of the gap, as in DSMath ($90\!\to\!98$, $+8$pp). At this budget, the endpoint mechanism adds little further lift: Qwen3 endpoints are capped at or near $100\%$, while the DSMath RL endpoint is slightly below SFT-ep3 ($95$ vs.\ $98$). Thus, in the easy regime, cross-surface robustness is limited mainly by saturation and the measurement ceiling rather than by a clearly separable endpoint mechanism. We therefore do not over-interpret within-family orderings, since with $n{=}40$ the one-standard-error scale is already about $\pm3$pp.

\paragraph{Hard regime (AIME): post-training expands the ceiling.}
On AIME 2025+2026, where substantial headroom remains, post-training
raises the large-$K$ ceiling over the base models. In Qwen3, sufficient
off-policy distillation already lifts $pass@224$ from $47\!\to\!77$ (4B)
and $55\!\to\!83$ (8B), and the released off-policy-plus-on-policy
endpoints raise the ceiling further to $90\%$ at both scales. Thus, in the hard regime, post-training is not only cheaper sampling of latent solutions; it can also expand reachable support.

\paragraph{Sample-efficiency vs.\ ceiling expansion.}
The mechanism-specific picture is more nuanced. In the DeepSeek-Math family, the released GRPO endpoint behaves mainly as sample-efficiency
reshaping: it improves pass@$1$ on the full set and AMC, but its large-$K$ values are flat or lower than our off-policy endpoint, and on AIME it
trails at both pass@$1$ and pass@$224$. By contrast, the Qwen3 endpoints
improve both pass@$1$ and the large-$K$ ceiling over sufficient off-policy
distillation. The headline pass@$1$ gain is still largely compression,
but the residual large-$K$ lift is genuine ceiling expansion. RL and
on-policy distillation are therefore not exchangeable post-training
mechanisms, even when both improve endpoint pass@$1$.

\paragraph{Scale saturates at the cross-surface ceiling, not at
$pass@1$.} After on-policy distillation, Qwen3-4B and Qwen3-8B endpoints nearly coincide on both $pass@1$ and $pass@224$ (AIME: $55.6/90$ vs.\
$57.2/90$; AMC: $87.6/100$ vs.\ $88.2/100$), suggesting that this
probe saturates by endpoint stage and that the residual scale gap is
mainly visible along the SFT trajectory.
\subsection{Multi-instance consistency}
\label{sec:mech:consistency}
Pass@$K$ asks whether \emph{any} sample succeeds; it does not say
whether success survives a surface change. We therefore measure,
for each source problem and probe class, whether all three variants
receive at least one correct sample at $K{=}16$ (\emph{all-3}),
whether any variant does (\emph{any-1}), and the conditional failure
rate among solved problems (\emph{fragile}). Table~\ref{tab:consistency}
reports T1 paraphrases; the per-problem T0/T1/T2 visualisation is in
Figure~\ref{fig:consistency} (Appendix~\ref{app:subset_tables}).

\begin{table}[!htb]
\centering
\footnotesize
\setlength{\tabcolsep}{6pt}
\begin{tabular}{@{}llrrr@{}}
\toprule
\textbf{Family} & \textbf{Stage} &
\textbf{all-3} & \textbf{any-1} & \textbf{fragile} \\
\midrule
\multirow{3}{*}{Qwen3-4B}
  & Base               & 26\% & 53\% & 51\% \\
  & SFT-ep3            & 58\% & 75\% & 23\% \\
  & \textbf{Endpoint}  & \textbf{84\%} & \textbf{89\%} & \textbf{6\%} \\
\midrule
\multirow{3}{*}{Qwen3-8B}
  & Base               & 33\% & 55\% & 40\% \\
  & SFT-ep3            & 65\% & 81\% & 20\% \\
  & \textbf{Endpoint}  & \textbf{87\%} & \textbf{90\%} & \textbf{3\%} \\
\midrule
\multirow{3}{*}{DSMath-7B}
  & Base               &  5\% & 22\% & 77\% \\
  & SFT-ep3            & 21\% & 46\% & 54\% \\
  & \textbf{Endpoint}  & \textit{20\%} & \textit{45\%} & \textit{56\%} \\
\bottomrule
\end{tabular}
\caption{\textbf{T1 multi-instance consistency (per source problem;
$3$ paraphrases, $K=16$ each).}
\textbf{all-3}: all variants solved; \textbf{any-1}: at least one
variant solved; \textbf{fragile} = (any-1 $-$ all-3) / any-1.}
\label{tab:consistency}
\end{table}

Consistency separates mechanisms: On T1, Qwen3 all-3 consistency increases monotonically from Base to
SFT-ep3 to Endpoint: $26\!\to\!58\!\to\!84\%$ for 4B and
$33\!\to\!65\!\to\!87\%$ for 8B, while fragility falls to
$6\%/3\%$. T0 and T2 show the same consolidation pattern
(Figure~\ref{fig:consistency}). DSMath is the contrast: its GRPO RL
endpoint is indistinguishable from SFT-ep3 ($20\%$ vs.\ $21\%$
all-3; $56\%$ vs.\ $54\%$ fragility). Together with the $pass@K$
inversion in \S\ref{sec:mech:ceiling}, this suggests that this
specific RL endpoint re-weights probability mass at $pass@1$ without
consolidating solved problems into surface-robust modes.

\subsection{Output-distribution signature: form vs.\ content sharpening}
\label{sec:mech:signature}

\begin{table}[!htb]
\centering
\scriptsize
\setlength{\tabcolsep}{3pt}
\begin{tabular}{@{}llrrrrr@{}}
\toprule
\textbf{Family} & \textbf{Stage} &
\textbf{maj} & \textbf{n$_\text{u}$} & \textbf{n$_\text{c}$} &
\textbf{emb} & \textbf{cv$_\text{len}$} \\
\midrule
\multirow{3}{*}{Qwen3-4B}
  & Base                 & 19.6\% & 26.29 &  3.50 & 0.312 & 1.447 \\
  & SFT-ep3              & 36.7\% &  8.36 &  1.00 & 0.123 & 0.390 \\
  & \textbf{Endpoint}    & \textbf{70.1\%} & \textbf{2.26} & \textbf{1.00} & \textbf{0.088} & \textbf{0.168} \\
\midrule
\multirow{3}{*}{Qwen3-8B}
  & Base                 & 19.4\% & 27.09 &  1.31 & 0.189 & 1.627 \\
  & SFT-ep3              & 48.4\% &  6.51 &  1.00 & 0.118 & 0.327 \\
  & \textbf{Endpoint}    & \textbf{69.5\%} & \textbf{2.08} & \textbf{1.00} & \textbf{0.091} & \textbf{0.147} \\
\midrule
\multirow{3}{*}{DSMath-7B}
  & Base                 &  7.4\% & 13.96 & 24.07 & 0.738 & 1.433 \\
  & SFT-ep3              & 11.4\% & 28.56 &  2.78 & 0.260 & 0.791 \\
  & \textbf{Endpoint}    & \textit{10.5\%} & \textit{9.76} & \textit{1.52} & \textit{0.224} & \textit{0.910} \\
\bottomrule
\end{tabular}
\caption{\textbf{Strategy-diversity descriptors on the T0 $K{=}48$
rollout pool} (\S\ref{sec:method:sampling}). Answers are extracted via
$\backslash\texttt{boxed}\{\cdot\}$ / \texttt{"Answer:"} /
\texttt{"final answer is"} patterns, then LaTeX-normalised.
\textbf{maj}: self-consistency majority share---fraction of the
$48$ samples giving the most common extracted
answer~\citep{wang2023self}.
\textbf{n$_{\text{u}}$}: distinct extracted answers per problem.
\textbf{n$_{\text{c}}$}: single-link clusters on MiniLM-L6
\emph{response} embeddings at cosine $\varepsilon{=}0.3$.
\textbf{emb}: mean pairwise embedding cosine distance.
\textbf{cv$_{\text{len}}$}: response-length coefficient of variation.
Italics flag the DSMath Endpoint, where GRPO RL leaves
$\text{maj}$ unchanged ($-0.9$pp) while contracting the answer
support $n_{\text{u}}$ by $-66\%$
(\S\ref{sec:mech:signature}).}
\label{tab:diversity}
\end{table}

\begin{figure}[!htb]
\centering
\includegraphics[width=\linewidth]{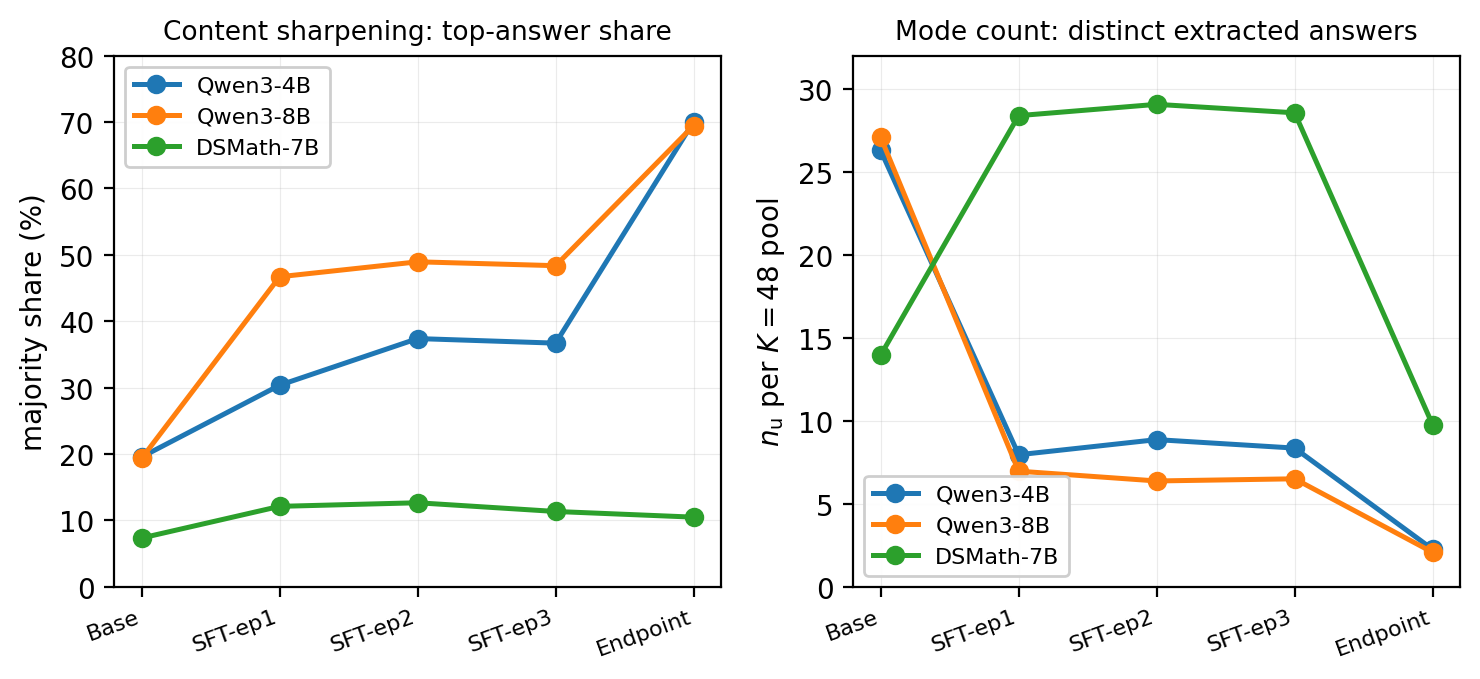}
\caption{\textbf{Content axis ($\text{maj}$, $n_{\text{u}}$)
across all checkpoints}
Answer extraction matches Table~\ref{tab:diversity}.}
\label{fig:sharpening}
\end{figure}

Pass@$K$ and consistency only reveal whether correct samples appear; they do not show how the rollout distribution changes. We therefore describe each T0 $K=48$ rollout pool along two axes. \textit{Form sharpening} measures whether reasoning traces collapse into similar surface realisations, using response-length variation and embedding-cluster diversity. \textit{Answer-content sharpening} measures whether the model concentrates on a small set of final answers, using self-consistency majority share and the number of distinct boxed answers. These axes are separable. Off-policy distillation consistently sharpens form, but its effect on answer content depends on the model family: Qwen3 contracts answer support, whereas DeepSeek-Math broadens it, likely because the hybrid SFT corpus introduces additional solution paths. The released Qwen3 on-policy-distilled endpoints sharpen both axes most strongly, reaching near-canonical reasoning traces and low answer diversity. By contrast, the DeepSeek-Math GRPO endpoint contracts the answer support relative to SFT but does not increase the majority-answer share (Fig.~\ref{fig:sharpening}). This is support contraction without modal consolidation. 

Together with the pass@$K$ and consistency results, this distributional view explains why mechanisms with similar endpoint gains need not be equivalent. The Qwen3 endpoints combine ceiling expansion, surface consistency, and answer-content consolidation. The DeepSeek-Math GRPO endpoint mainly prunes the SFT answer support without producing a higher large-$K$ ceiling or stronger surface consistency in our setting.

\section{Cross-lingual generalisation}
\label{sec:multilingual}
Multilingual math reasoning has been evaluated on fixed translation
sets such as Multilingual Grade School Math (MGSM)~\citep{shi2210language}. We extend this idea to
competition-style source problems and pool attempts within the same
source problem. T3 (Section~\ref{sec:method:probes}) covers $5$ translations per
source (Chinese, Spanish, Arabic, Urdu, Swahili).
For this section we extend T3 with $3$ additional languages chosen
to broaden script and language-family coverage---Swedish,
Icelandic, Tamil---using the same translation protocol and $K{=}16$
per (source, language). Combined with $16$ T0 samples per source as
the English anchor, this yields a $9$-language cross-section over
the same $100$ source problems for all $9$ models. Languages are
grouped by Qwen3-4B Base $pass@16$:
\textit{stronger non-en} $=$ \{zh, es, sv, ar\}, the four
non-English languages with the highest Qwen3-4B Base $pass@16$;
\textit{weaker non-en} $=$ \{is, ta, ur, sw\}, the remaining four.

Three readings across the $9$ (family, role) cells of
the multilingual table (full numbers and per-family bar charts
in Appendix~\ref{app:multilingual_figs},
Table~\ref{tab:multilingual}): \textbf{(F1)} the stronger
non-English group $\{$zh, es, sv, ar$\}$ beats the weaker group
$\{$is, ta, ur, sw$\}$ in every cell by $+3$ to $+18$pp on mean
$pass@16$, with the tier preserved by every mechanism; \textbf{(F2)}
English-dominant SFT lifts every non-English language by $+4$ to
$+49$pp at SFT-ep3 across all $24$ non-English cells (largest:
Icelandic on Qwen3-8B, $23{\to}72$)---even Swahili moves up
despite no direct supervision (\S\ref{sec:method:sft}); \textbf{(F3)}
Swahili is a floor that no mechanism removes, trailing the
next-worst language by $24/17/10$pp at the Qwen3-4B/8B/DSMath-7B
endpoints, and regressing $9{\to}4$ on DSMath-7B from SFT-ep3 to
RL endpoint (the largest \emph{relative} single-language
regression in the table at $-56\%$; DSMath es also drops
$33{\to}27$, $-6$pp absolute)---a deficit fixed at Base and
preserved or amplified by every mechanism, suggesting a
pretraining-corpus rather than post-training property (visualised
across $6$ models in Appendix
Fig.~\ref{fig:multilingual_imbalance}). Pooling all
$9$ languages ($16$ samples each, $144$ per source) reproduces the
same pattern as \S\ref{sec:mech:ceiling}: on DSMath, GRPO RL leads
at small $K$ (sharpening), and SFT-ep3 matches or exceeds it from
$K{\approx}16$ onward. Qwen3 endpoints remain monotone-above
SFT-ep3 throughout (Appendix Fig.~\ref{fig:passK_9lang}).

\section{Memorisation}
We report a controlled-overfit memorisation audit on $3$ families
$\times$ $3$ epochs ($0$, $10$, $20$) $\times$ $3$ probe modalities
(behavioural prefix-completion, behavioural forbidden-CoT,
forward-pass membership-inference attack (MIA)~\citep{shi2024detecting}). The overfit corpus is
a $200$-item subset of each family's canonical training source
with an MD5-disjoint $200$-item held-out set; $10$/$20$ epochs
deliver an order of magnitude more per-item gradient passes than
a typical SFT trajectory, designed to maximise detectability.
Under these conditions, none of the three modalities yields a
strong, reliable memorisation signal: behavioural $\Delta$'s are
noisy across families and probe types (we omit them from the
quantitative claims), and forward-pass MIA reports area under
the receiver-operating-characteristic curve (ROC AUC) bounded
within $[.51, .65]$ across the overfit grid
(Appendix~Table~\ref{tab:mia}), below the stronger
pretraining-data detection signals reported in prior work
(e.g., Min-K\% Prob reaching AUC $0.88$ for copyrighted-book
detection~\citep{shi2024detecting}). Read as a \emph{sensitivity ceiling},
this audit calibrates what current methods can resolve about
SFT-specific membership and cautions against interpreting modest
MIA AUCs as direct evidence of SFT memorisation. The full
protocol is in Appendix~\ref{sec:memorisation}.


\section{Discussion}

Our results suggest that mathematical post-training is best understood as distribution shaping over a difficulty-dependent capability landscape. Endpoint pass@$1$ conflates several effects: cheaper sampling of already reachable solutions, expansion of the large-$K$ reachable ceiling, robustness to surface variation, and possible memorisation. These effects can move differently across mechanisms and therefore require trajectory-aware diagnosis.

The compression account is real, but regime-bound. On easier AMC problems, the cross-surface large-$K$ ceiling is near saturation, so post-training mainly improves sample efficiency: correct solutions become cheaper to sample while mechanisms converge near the same ceiling. On harder AIME problems, however, base models leave substantial headroom, and post-training expands the reachable ceiling. Crucially, this expansion already appears under our sufficiently trained off-policy distillation trajectories, making offline distillation a genuine capability-expanding post-training path rather than a weak SFT baseline. Released endpoints then provide mechanism-relative evidence: Qwen3 off-policy-plus-on-policy endpoints add further large-$K$ lift, whereas the DeepSeek-Math GRPO endpoint sharpens small-$K$ behaviour but does not exceed sufficient off-policy distillation at large $K$ in our comparison.

Robustness and distributional diagnostics further show that mechanisms are not interchangeable. Qwen3 endpoints consolidate solved problems across paraphrases and numerical isomorphisms, while the DeepSeek-Math GRPO endpoint remains close to its SFT predecessor in multi-instance consistency. Form sharpening and answer-content sharpening can also diverge, so more uniform reasoning traces should not be equated with more reliable reasoning unless correctness and cross-surface consistency improve as well. Cross-lingually, English-dominant distillation improves non-English reasoning but preserves stronger-versus-weaker language gaps, suggesting that post-training does not erase inherited coverage asymmetries. Finally, our controlled-overfit audit calibrates how much memorisation current membership probes can detect at all: the weak signals we measure bound the probes, not the phenomenon.

\section{Conclusion}
We compare three mathematical post-training mechanisms---off-policy distillation, off-policy-plus-on-policy distillation, and off-policy-plus-GRPO RL---under a shared cross-surface diagnostic framework. Using cross-surface pass@$K$, consistency, distribution-shape, multilingual, and verified SFT-membership probes, we separate sample-cost compression from ceiling expansion, robustness, and memorisation sensitivity.

Overall, compression is one regime of post-training, not a universal explanation of reasoning gains. Easy problems mainly expose sampling-cost compression, while hard problems show that sufficient post-training---already off-policy distillation---can expand the reachable ceiling over the base. Evaluating reasoning progress therefore requires surface-aware, trajectory-aware, and membership-controlled diagnostics beyond endpoint pass@$1$.

\section*{Limitations}
Our comparisons are mechanism-relative rather than fully causal: the studied post-training paths differ in historical pipeline, and we isolate no single algorithmic factor.

We cover nine languages, but translation quality, cultural specificity, and benchmark familiarity vary across them. The memorisation study calibrates detection sensitivity on our own controlled SFT data; it does not audit contamination from pretraining or from the data behind released endpoints.

\section*{Ethical Statement}
This work uses publicly available competition mathematics
problems (AIME 2025/2026, AMC 2023) and public open-source
training corpora (OpenR1-Math, NuminaMath-CoT). No human
subjects, personally identifiable information, or sensitive
content are involved. All models analysed are openly released
under their respective licenses, and our SFT training data is
filtered and curated from public corpora rather than collected
from new sources. The full reproducibility package---problem pools and surface
transforms, rollouts, analysis intermediates, and the analysis
code---accompanies this submission as the code and data archive,
so every reported number can be independently verified.

\bibliography{refs}

\appendix

\section{Formal definition of cross-surface pass@$K$}
\label{app:formal_passk}

For each source problem $s$ and a subset of transforms
$\mathcal{T}'\!\subseteq\!\{T_0,T_1,T_2,T_3\}$, let $N$ be the
size of the pool of all rollouts drawn from variants of $s$
under transforms in $\mathcal{T}'$, and let $n$ be the count of
those rollouts scored correct by \textsf{math\_verify}. The
unbiased pass@$K$ estimator \citep{chen2021evaluating} is
\begin{equation*}
\mathrm{pass@}K(s)
=
\begin{cases}
0, & n=0,\\[2pt]
1, & N-n<K,\\[4pt]
1 - \binom{N-n}{K}\!\big/\!\binom{N}{K}, & \text{else.}
\end{cases}
\end{equation*}
We report the source-level mean on a subset
$\mathcal{S}'\!\subseteq\!\{s_1,\dots,s_{100}\}$, denoted
$\mathrm{pass@}K(\mathcal{S}')$. Every reported cell has
$K\le N$. The pool behind each column is given by the
\emph{probe set} row of each pass@$K$ table, and $\mathcal{S}'$ by
its caption. Two places do not fit this formalism and state their own
convention instead: the multilingual table draws its \texttt{en}
column from the first $16$ of the $48$ T0 samples and each
non-English column from the $16$ T3 samples of that language, and the
$9$-language pooled figures combine those same $16$ T0 samples with
eight T3 languages.

\section{Training details}
\label{app:training}

\begin{figure*}[!tb]
\centering
\begin{subfigure}{0.32\textwidth}
\centering
\includegraphics[width=\linewidth]{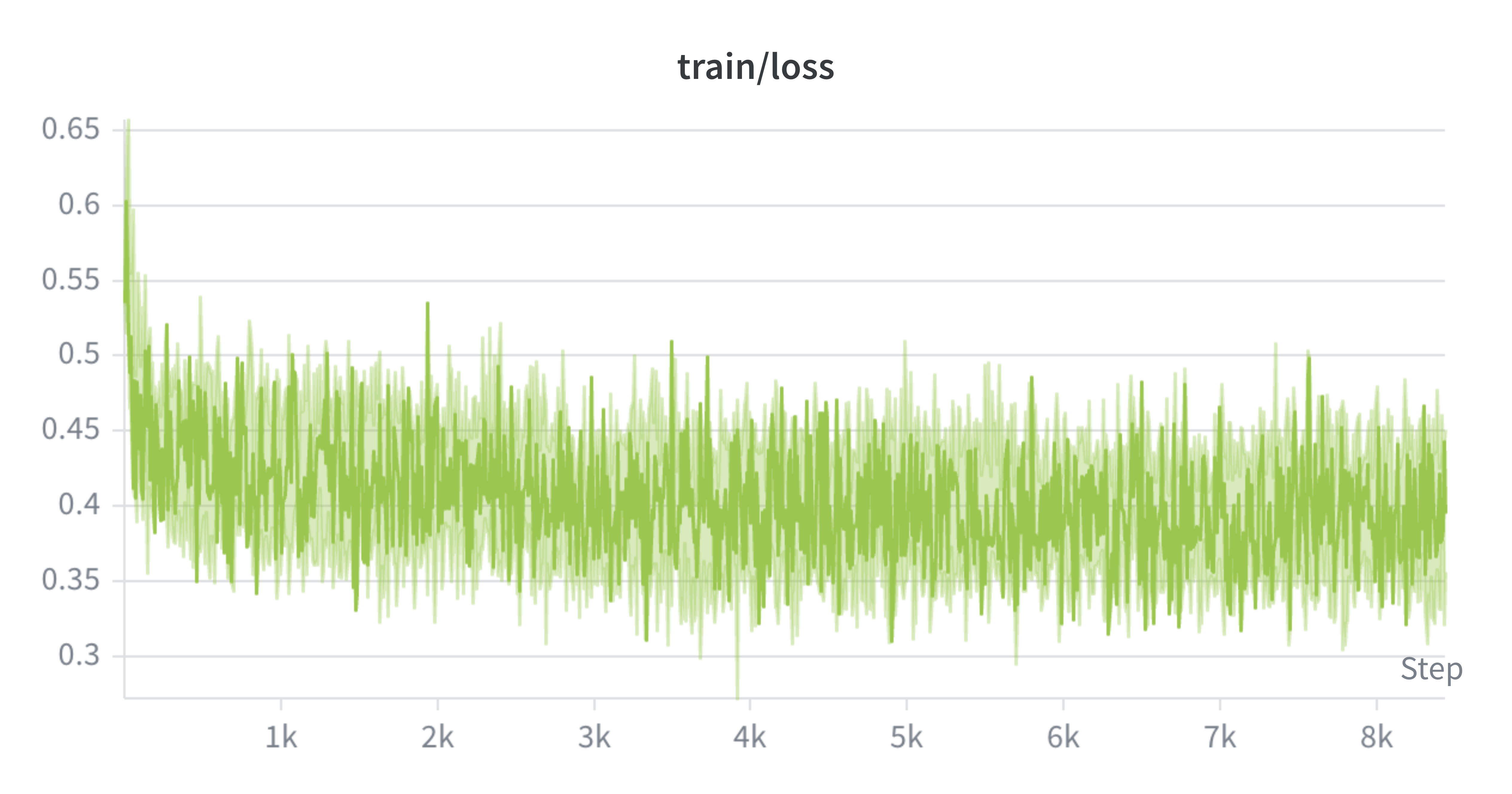}
\caption{Qwen3-4B \texttt{train/loss}}
\label{fig:loss_4b}
\end{subfigure}\hfill
\begin{subfigure}{0.32\textwidth}
\centering
\includegraphics[width=\linewidth]{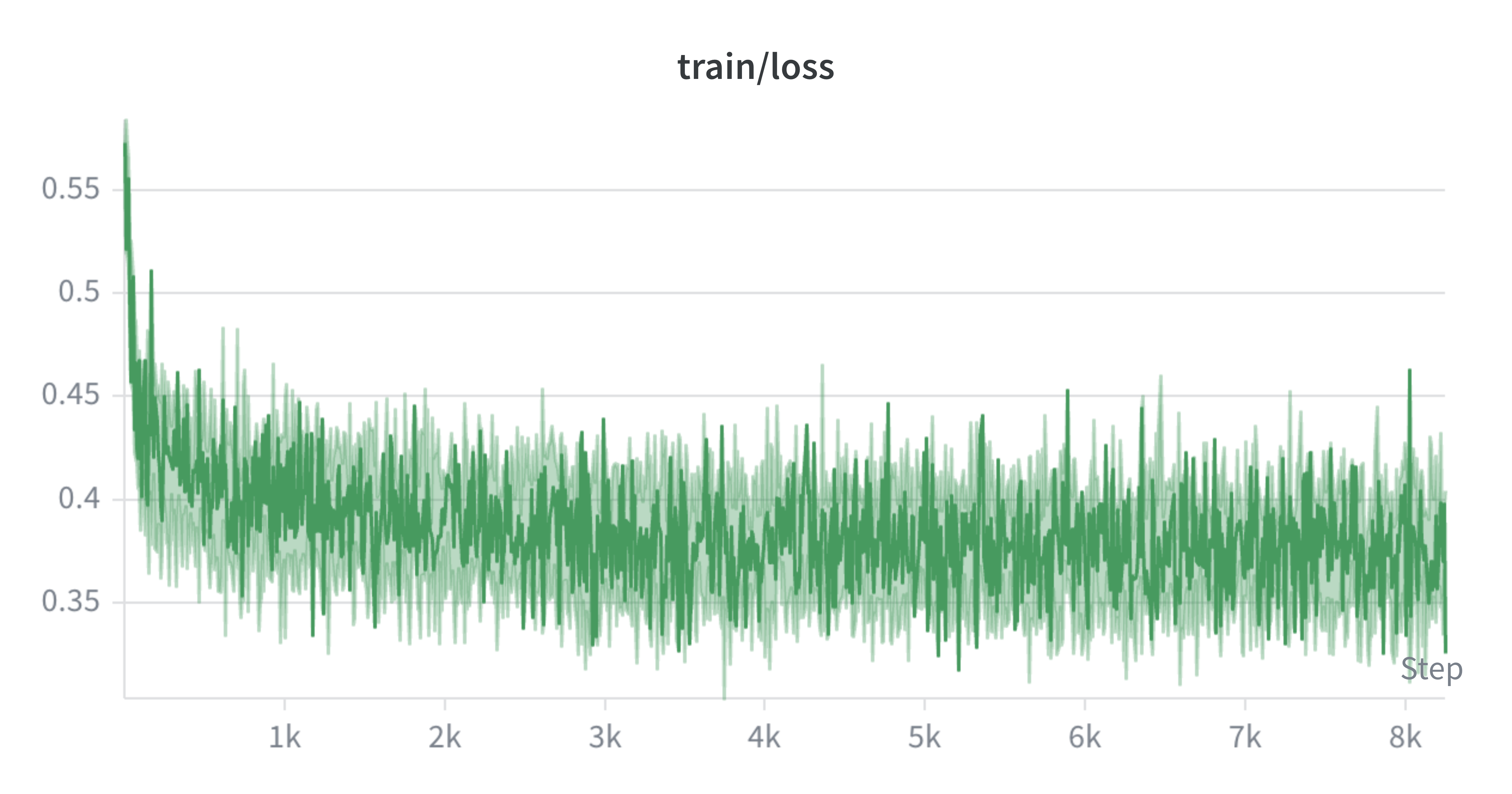}
\caption{Qwen3-8B \texttt{train/loss}}
\label{fig:loss_8b}
\end{subfigure}\hfill
\begin{subfigure}{0.32\textwidth}
\centering
\includegraphics[width=\linewidth]{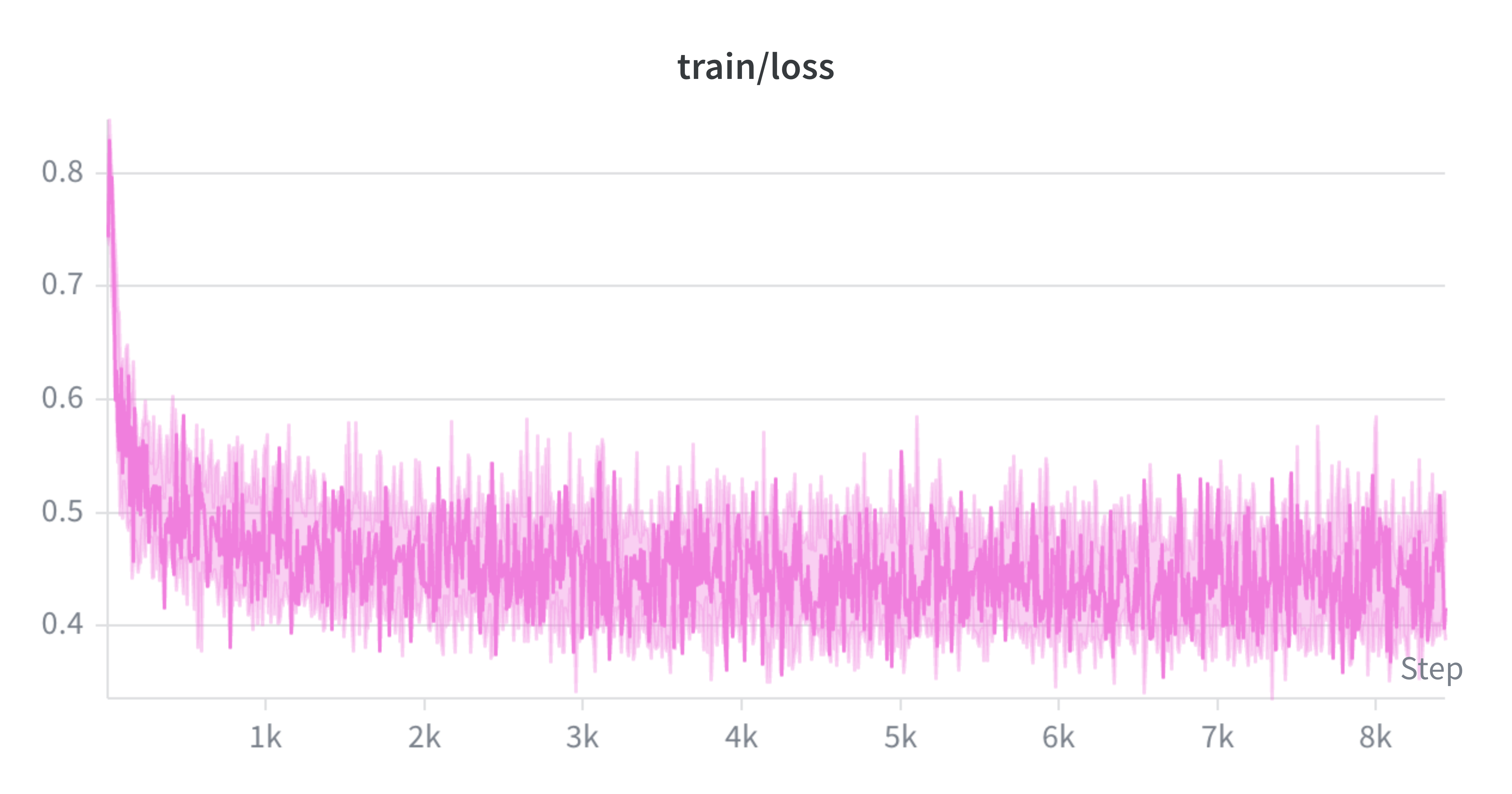}
\caption{DSMath-7B \texttt{train/loss}}
\label{fig:loss_7b}
\end{subfigure}
\\[4pt]
\begin{subfigure}{0.32\textwidth}
\centering
\includegraphics[width=\linewidth]{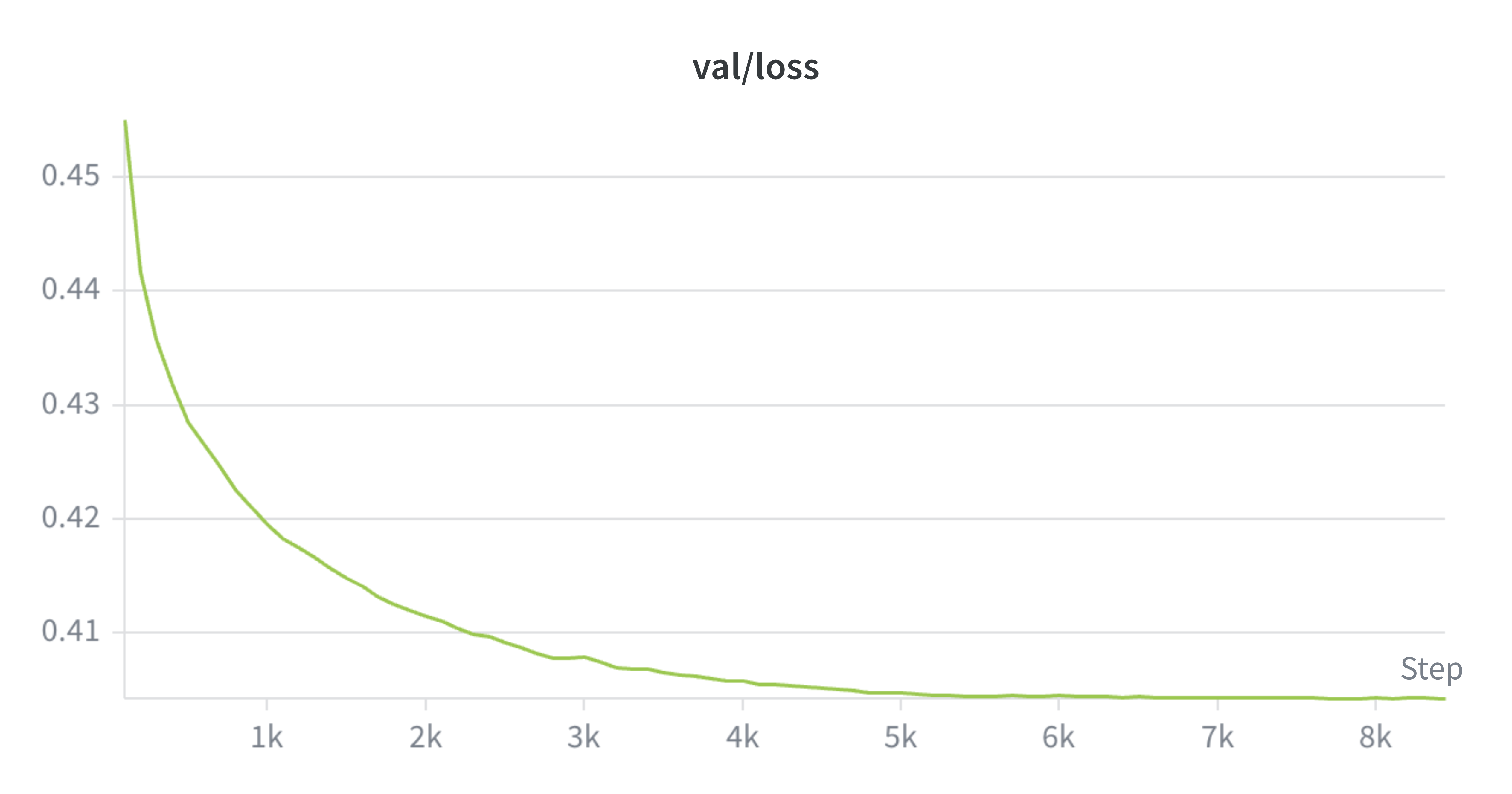}
\caption{Qwen3-4B \texttt{val/loss}}
\label{fig:val_loss_4b}
\end{subfigure}\hfill
\begin{subfigure}{0.32\textwidth}
\centering
\includegraphics[width=\linewidth]{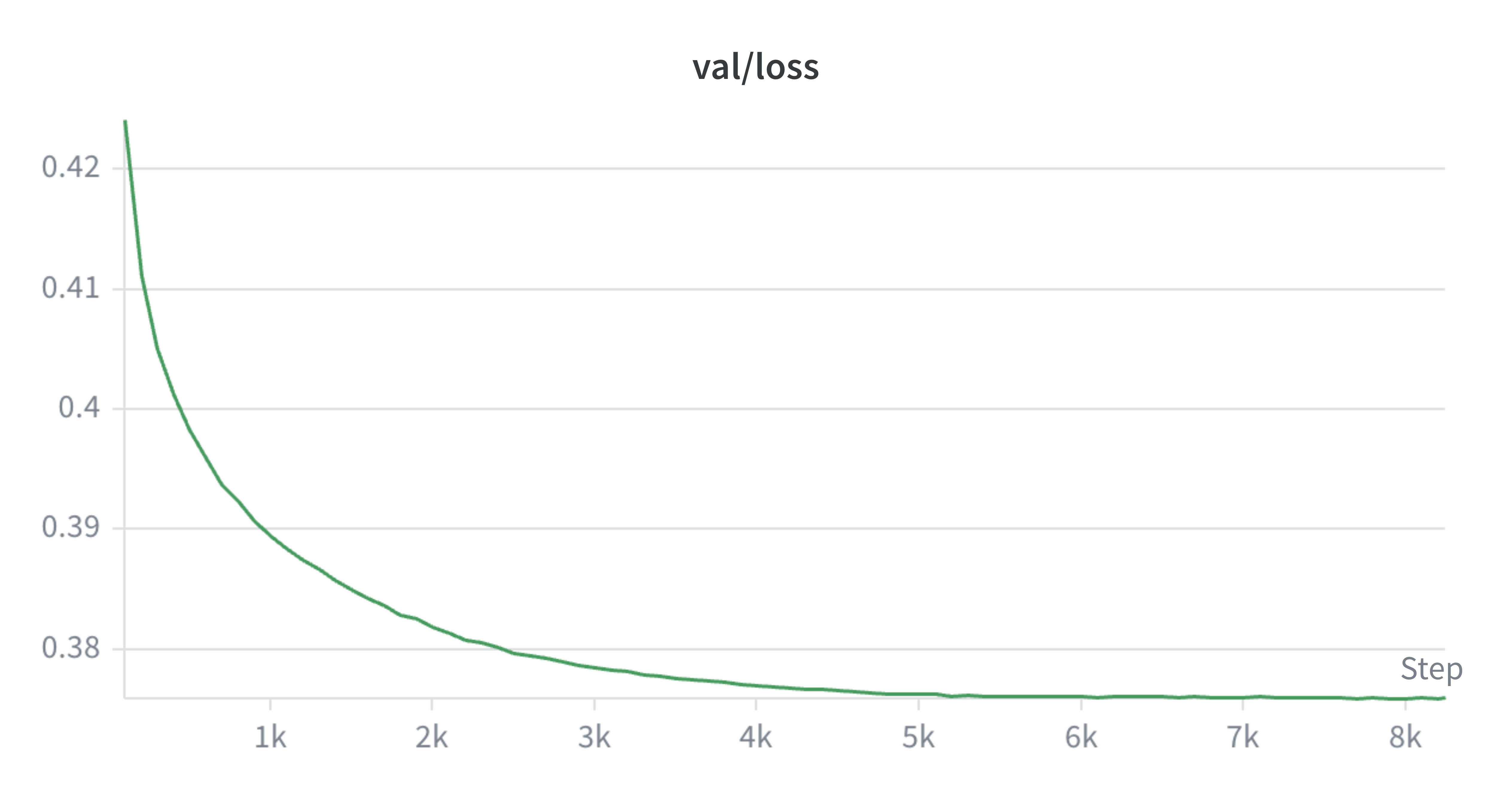}
\caption{Qwen3-8B \texttt{val/loss}}
\label{fig:val_loss_8b}
\end{subfigure}\hfill
\begin{subfigure}{0.32\textwidth}
\centering
\includegraphics[width=\linewidth]{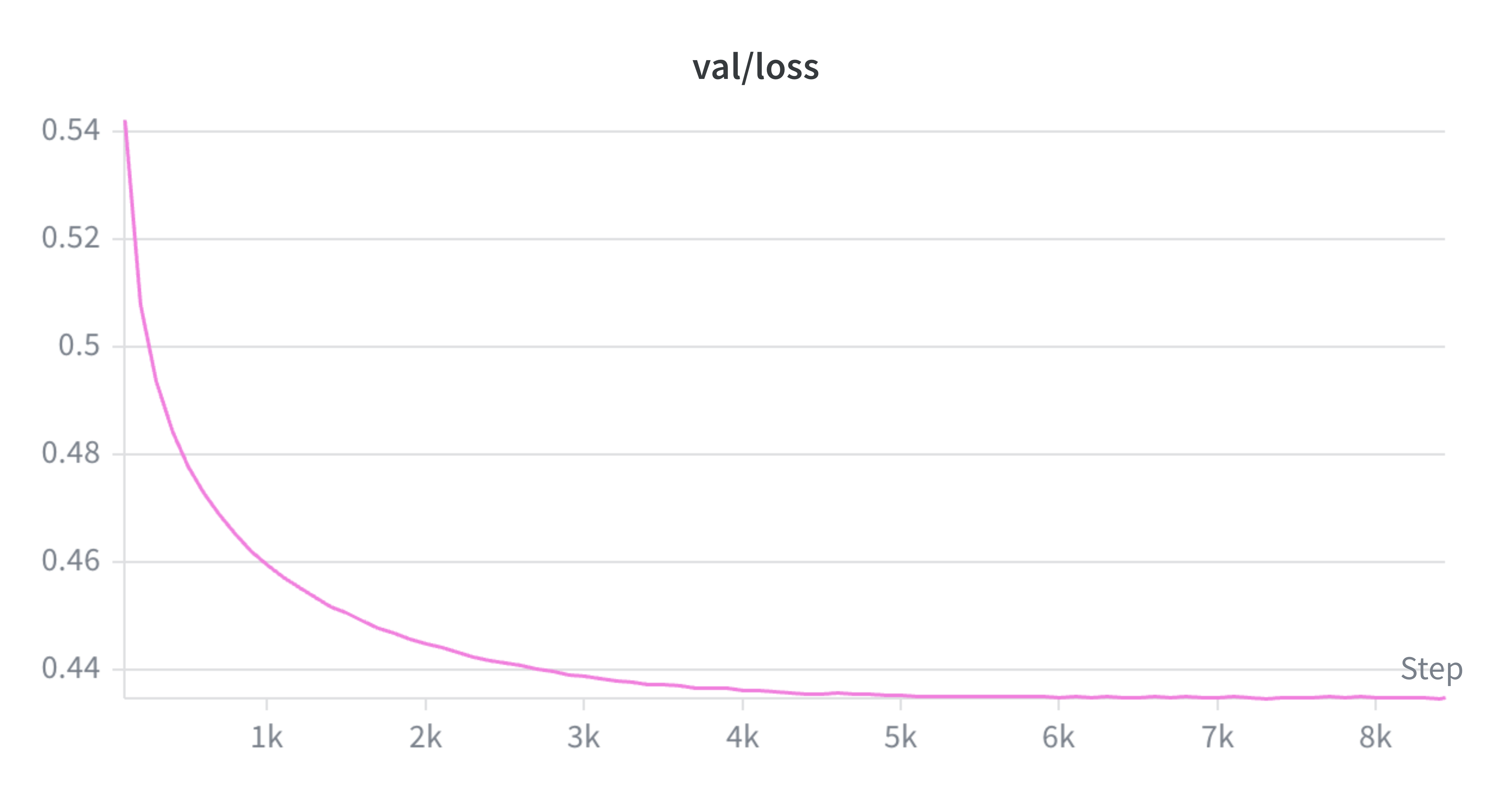}
\caption{DSMath-7B \texttt{val/loss}}
\label{fig:val_loss_7b}
\end{subfigure}
\caption{Training (top row) and validation (bottom row) loss across
three epochs of off-policy distillation.}
\label{fig:loss_curves}
\end{figure*}

\begin{figure*}[!tb]
\centering
\begin{minipage}{0.32\textwidth}
\includegraphics[width=\linewidth]{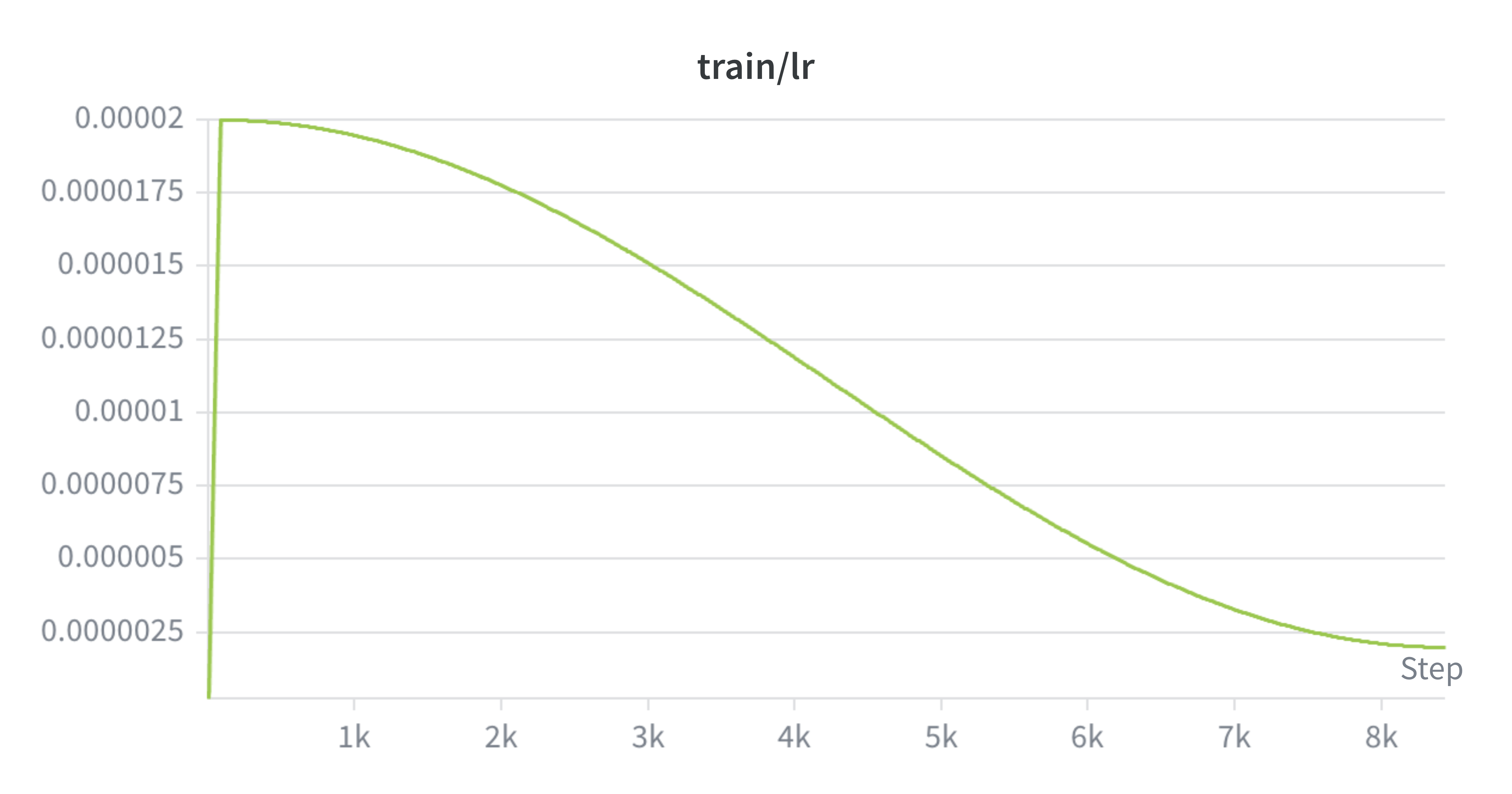}
\end{minipage}\hfill
\begin{minipage}{0.32\textwidth}
\includegraphics[width=\linewidth]{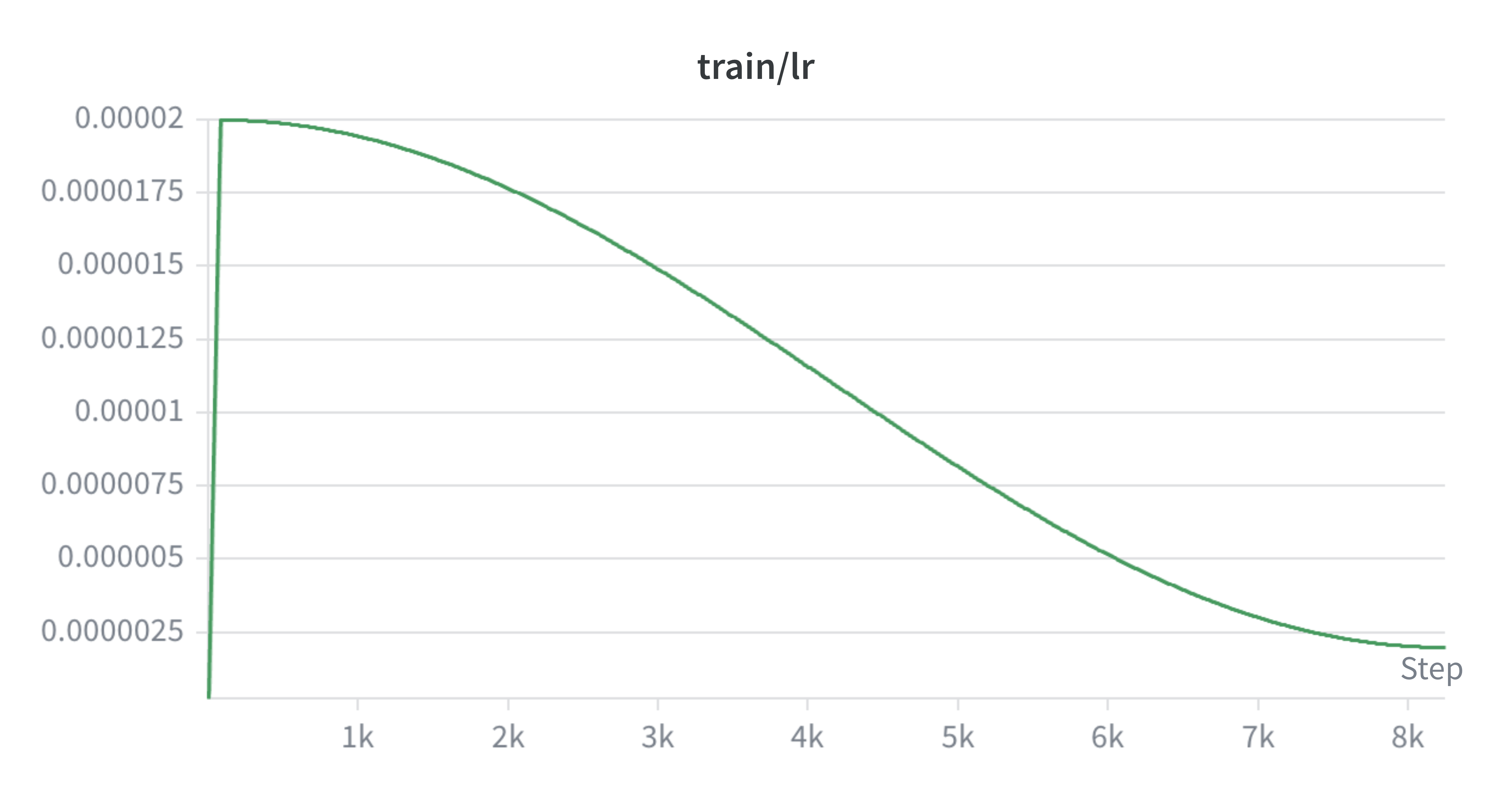}
\end{minipage}\hfill
\begin{minipage}{0.32\textwidth}
\includegraphics[width=\linewidth]{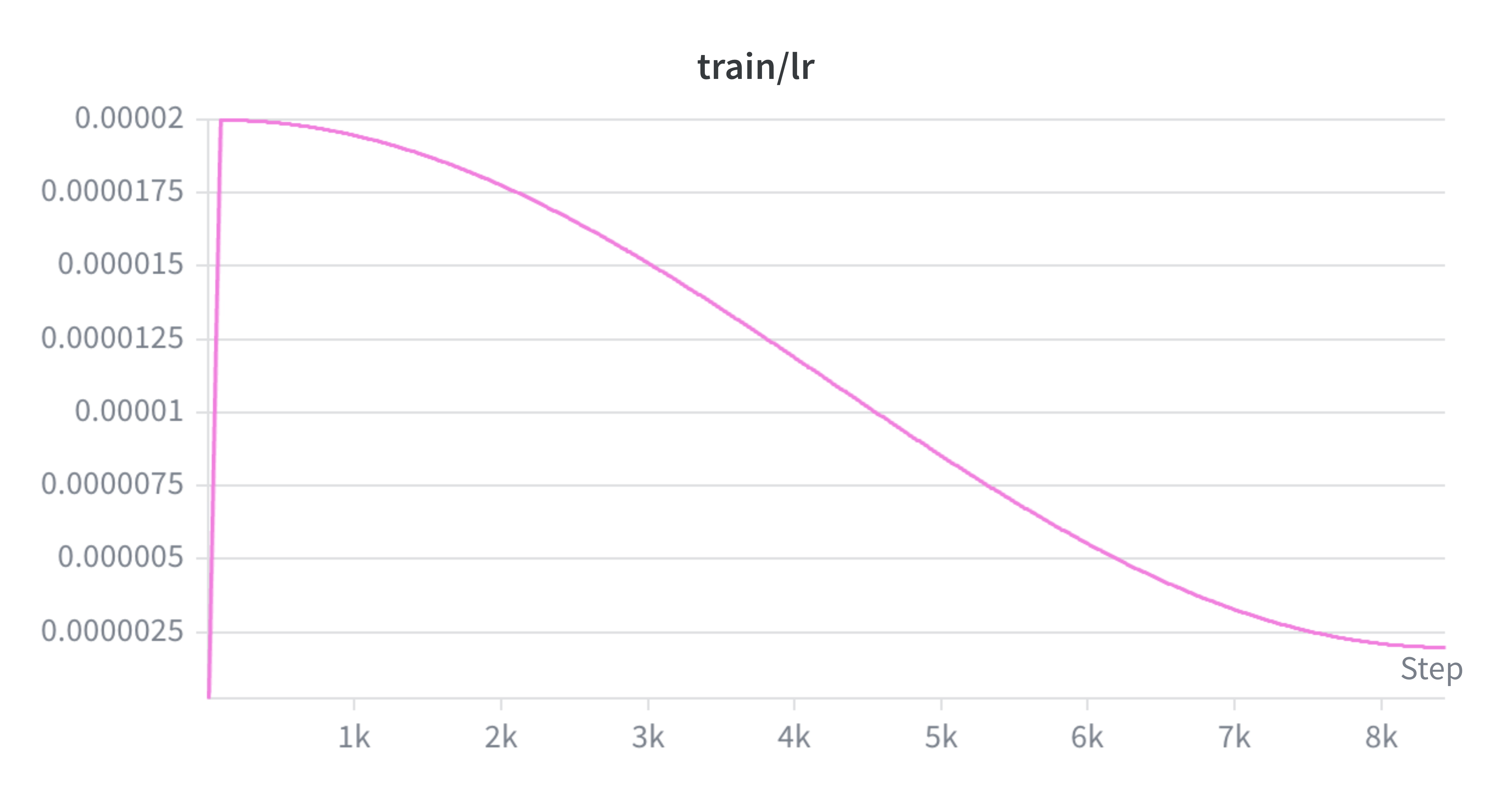}
\end{minipage}
\caption{Learning-rate schedule for the SFT runs. Left to right:
\texttt{Qwen3-4B-SFT-Math-45k}, \texttt{Qwen3-8B-SFT-Math-90k},
and \texttt{DSMath-7B-SFT-hybrid}}
\label{fig:lr_curves}
\end{figure*}

\begin{figure*}[!tb]
\centering
\begin{minipage}{0.32\textwidth}
\includegraphics[width=\linewidth]{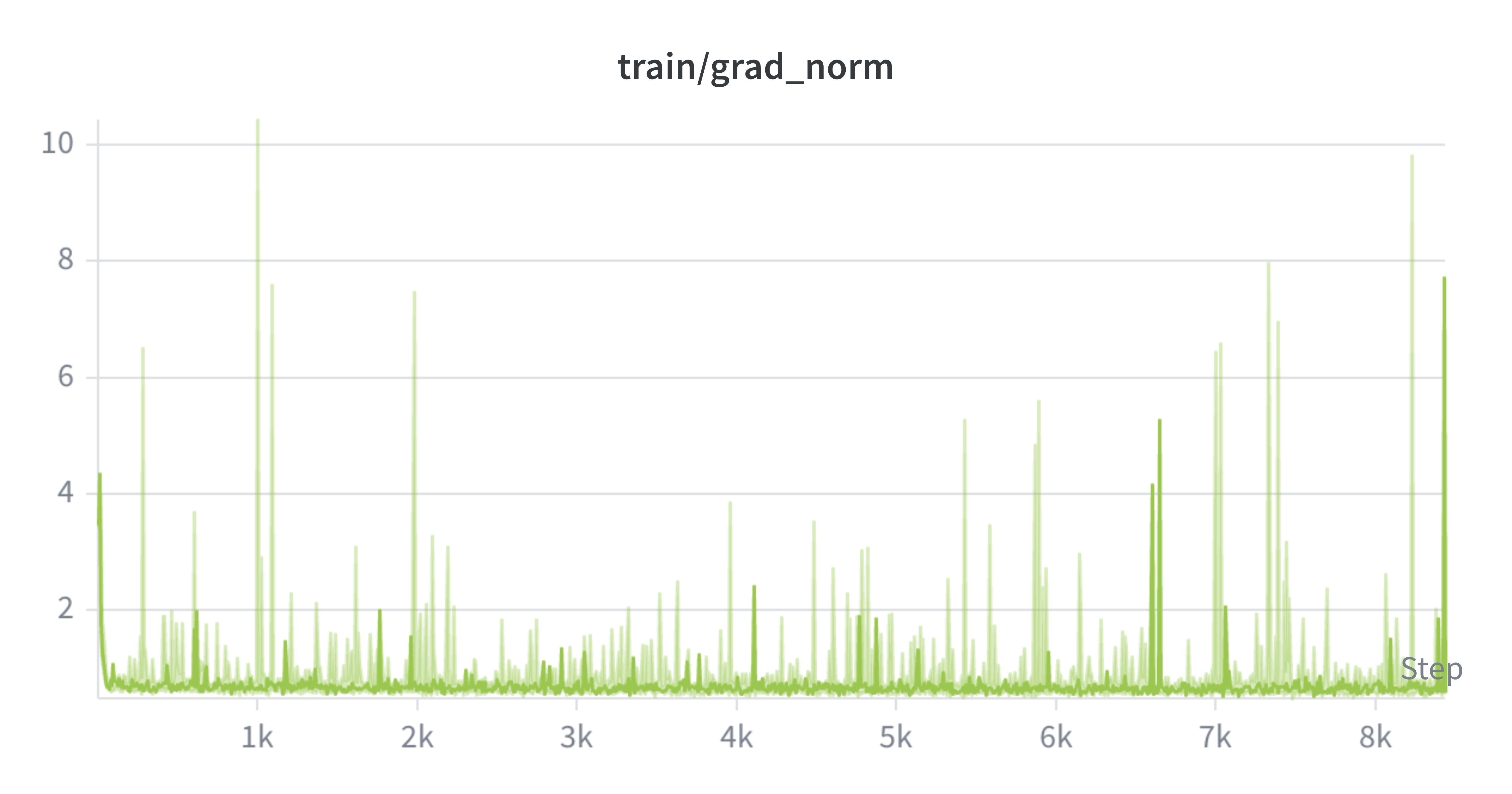}
\end{minipage}\hfill
\begin{minipage}{0.32\textwidth}
\includegraphics[width=\linewidth]{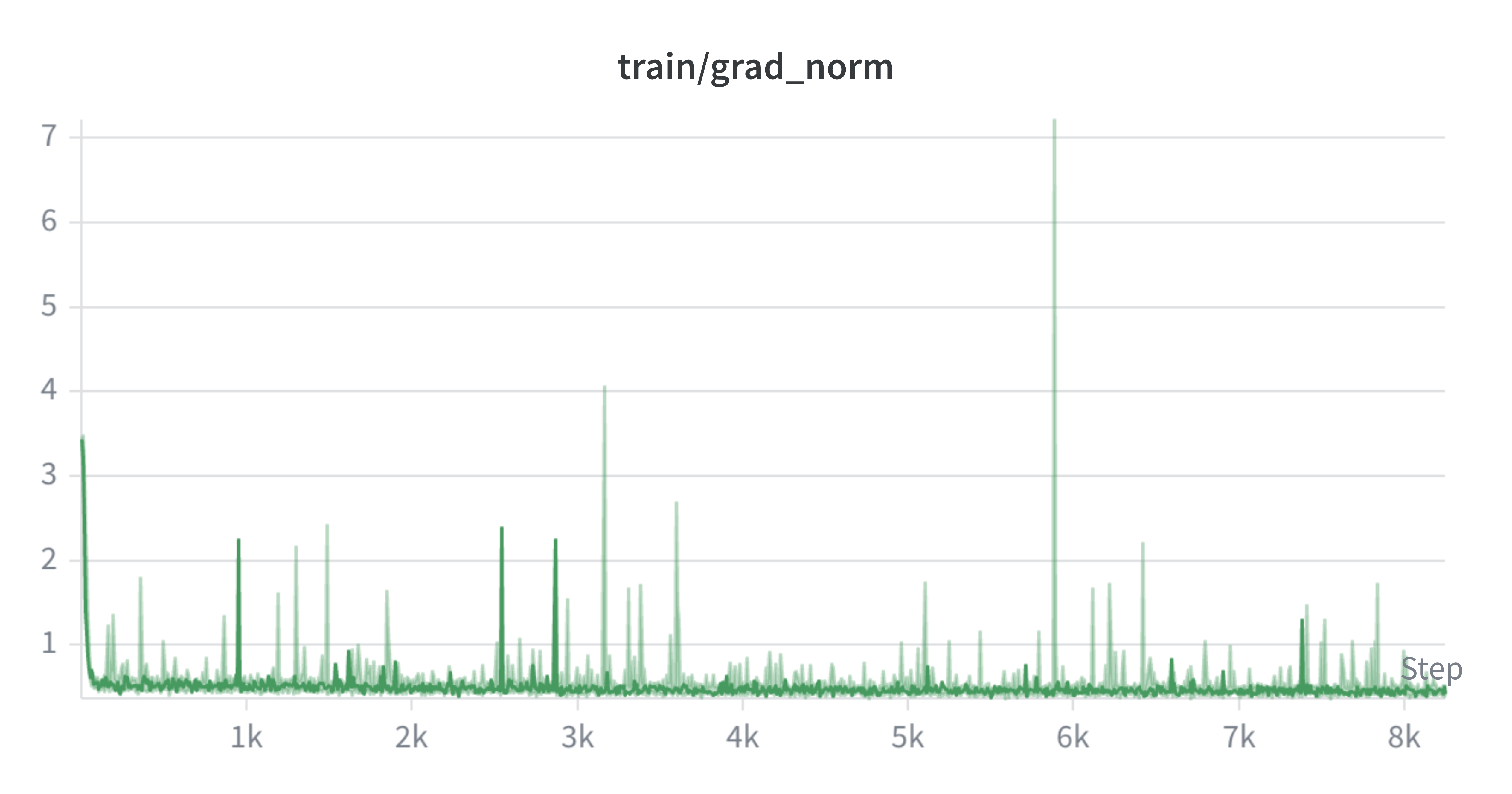}
\end{minipage}\hfill
\begin{minipage}{0.32\textwidth}
\includegraphics[width=\linewidth]{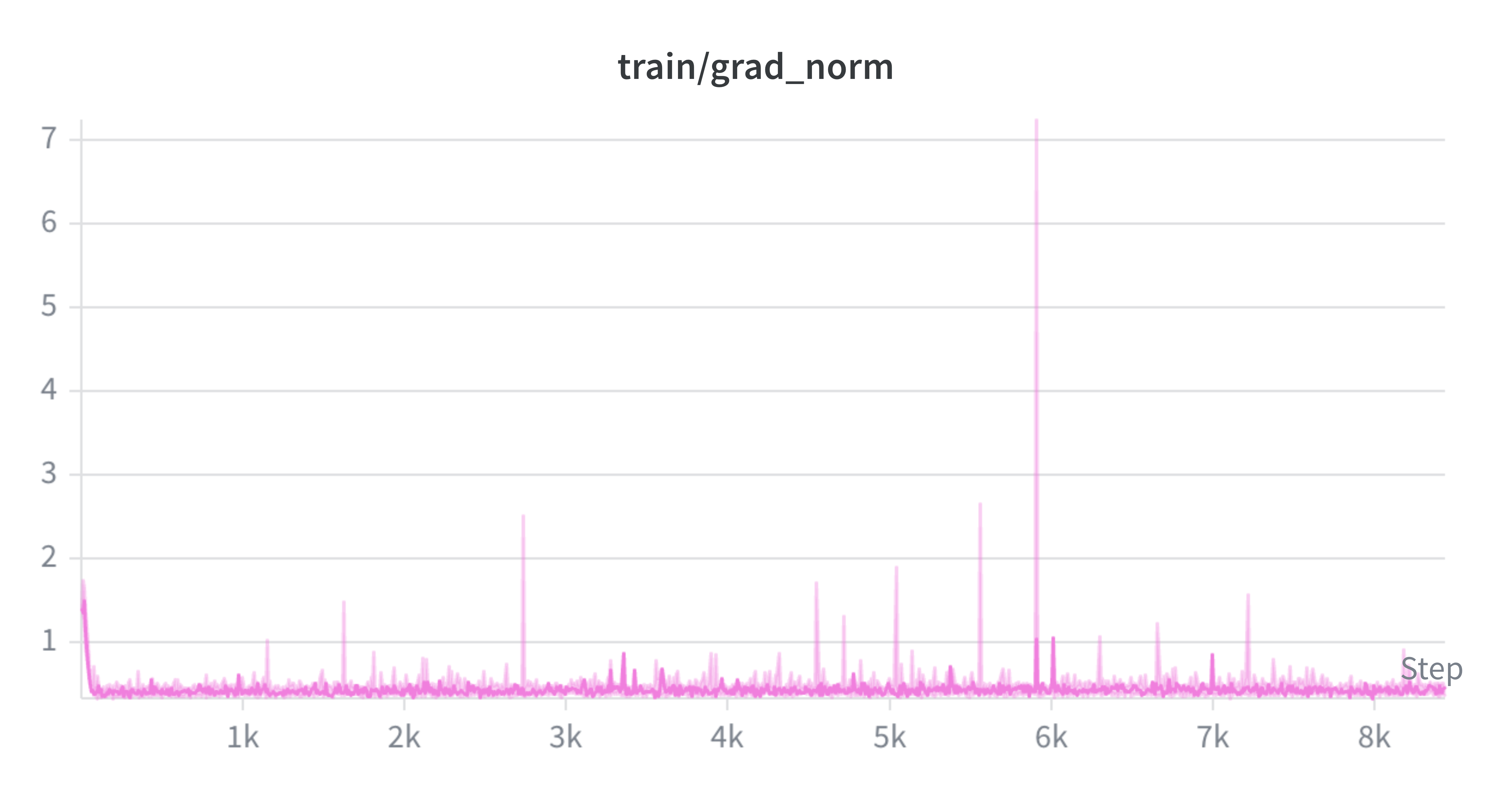}
\end{minipage}
\caption{Gradient norm for the SFT runs (clip threshold $1.0$, see
Table~\ref{tab:hyper}). No divergence over three epochs.}
\label{fig:grad_norm}
\end{figure*}

This appendix reports the full hyperparameter list
(Table~\ref{tab:hyper}), training/validation loss curves
(Figure~\ref{fig:loss_curves}), and per-step learning-rate
(Figure~\ref{fig:lr_curves}) and gradient-norm
(Figure~\ref{fig:grad_norm}) trajectories for the representative SFT
runs, confirming that the cosine schedule listed in
Table~\ref{tab:hyper} was executed as specified and that no run
diverged.
Figure~\ref{fig:lr_curves}: peak $2\times10^{-5}$ is reached after
$\sim 1\%$ of total steps (linear warm-up) and decays smoothly to
$2\times10^{-6}$, exactly $0.1\times$ peak as set by
\texttt{min\_lr\_ratio}.
Figure~\ref{fig:grad_norm}: gradient norm stays bounded throughout
training, with no divergence or loss explosion.

\begin{table*}[!tb]
\centering
\small
\setlength{\tabcolsep}{10pt}
\begin{tabular}{ll}
\toprule
\textbf{Hyperparameter} & \textbf{Value} \\
\midrule
Optimiser                       & AdamW, $(\beta_1, \beta_2) = (0.9, 0.999)$ \\
Peak learning rate              & $2 \times 10^{-5}$ \\
LR schedule                     & Cosine, min ratio $0.1$ \\
Warm-up                         & $1\%$ of total steps \\
Weight decay                    & $0.1$ \\
Gradient clipping               & $1.0$ \\
Precision                       & bfloat16 (FSDP, activation ckpt; activation
                                  offload on the 8B and 7B runs only) \\
Max sequence length             & $32{,}768$ (Qwen3) / $4{,}096$ (DSMath, native) \\
Micro-batch per GPU             & $1$ \\
Global batch (4B)               & $16$ \\
Global batch (7B / 8B)          & $32$ \\
Epochs                          & $3$ \\
Total tokens (Qwen3-4B)         & $\approx 0.80$\,B \\
Total tokens (Qwen3-8B)         & $\approx 1.56$\,B \\
Total tokens (DSMath-7B)        & $\approx 0.34$\,B \\
\bottomrule
\end{tabular}
\caption{SFT training hyperparameters.}
\label{tab:hyper}
\end{table*}

\section{Cross-mechanism analysis}
\label{app:subset_tables}

\begin{figure*}[!tb]
\centering
\begin{subfigure}[t]{0.49\linewidth}
  \centering
  \includegraphics[width=\linewidth]{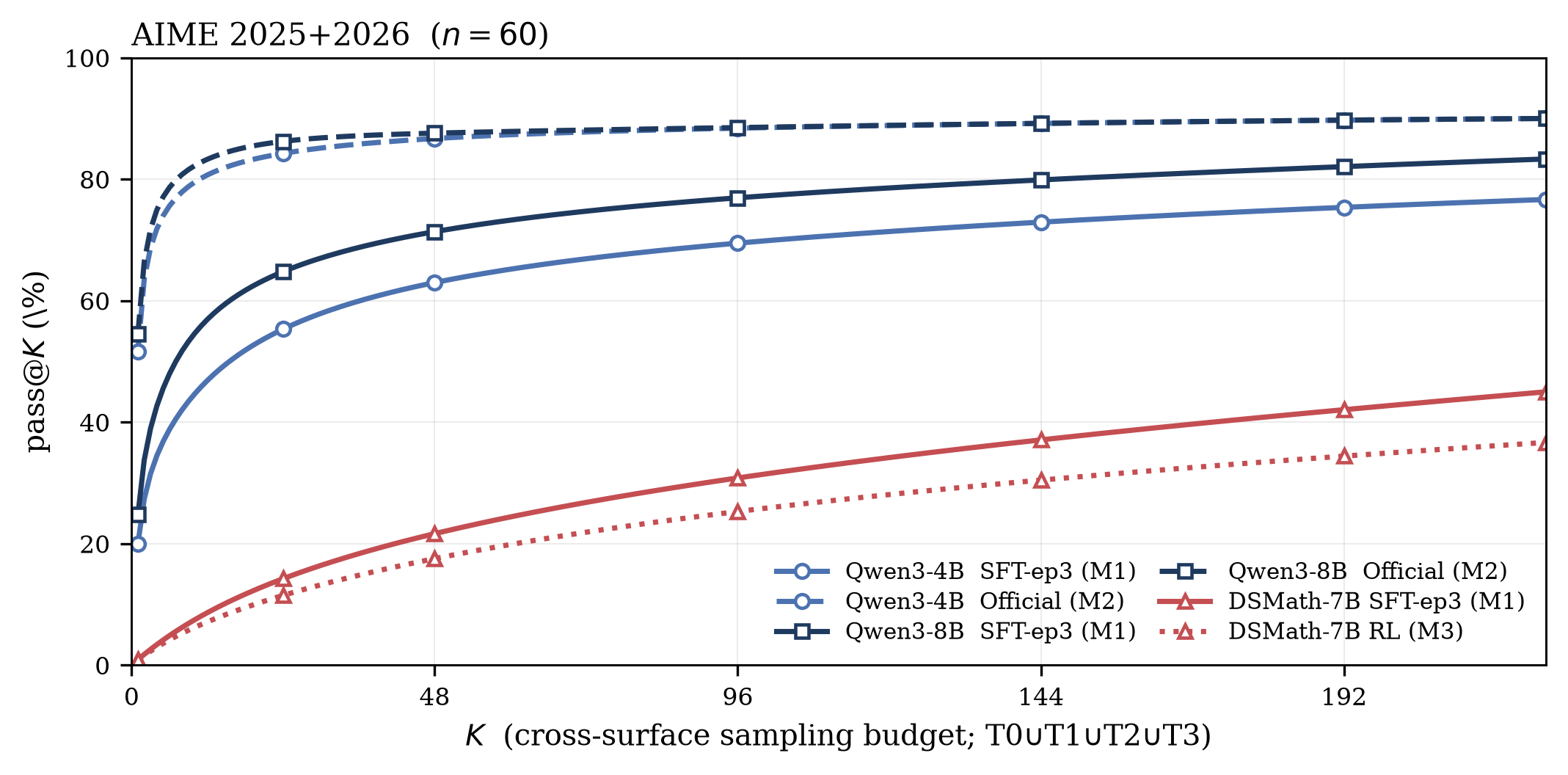}
  \caption{AIME 2025+2026 ($n{=}60$).}
  \label{fig:passK_combined_aime}
\end{subfigure}\hfill
\begin{subfigure}[t]{0.49\linewidth}
  \centering
  \includegraphics[width=\linewidth]{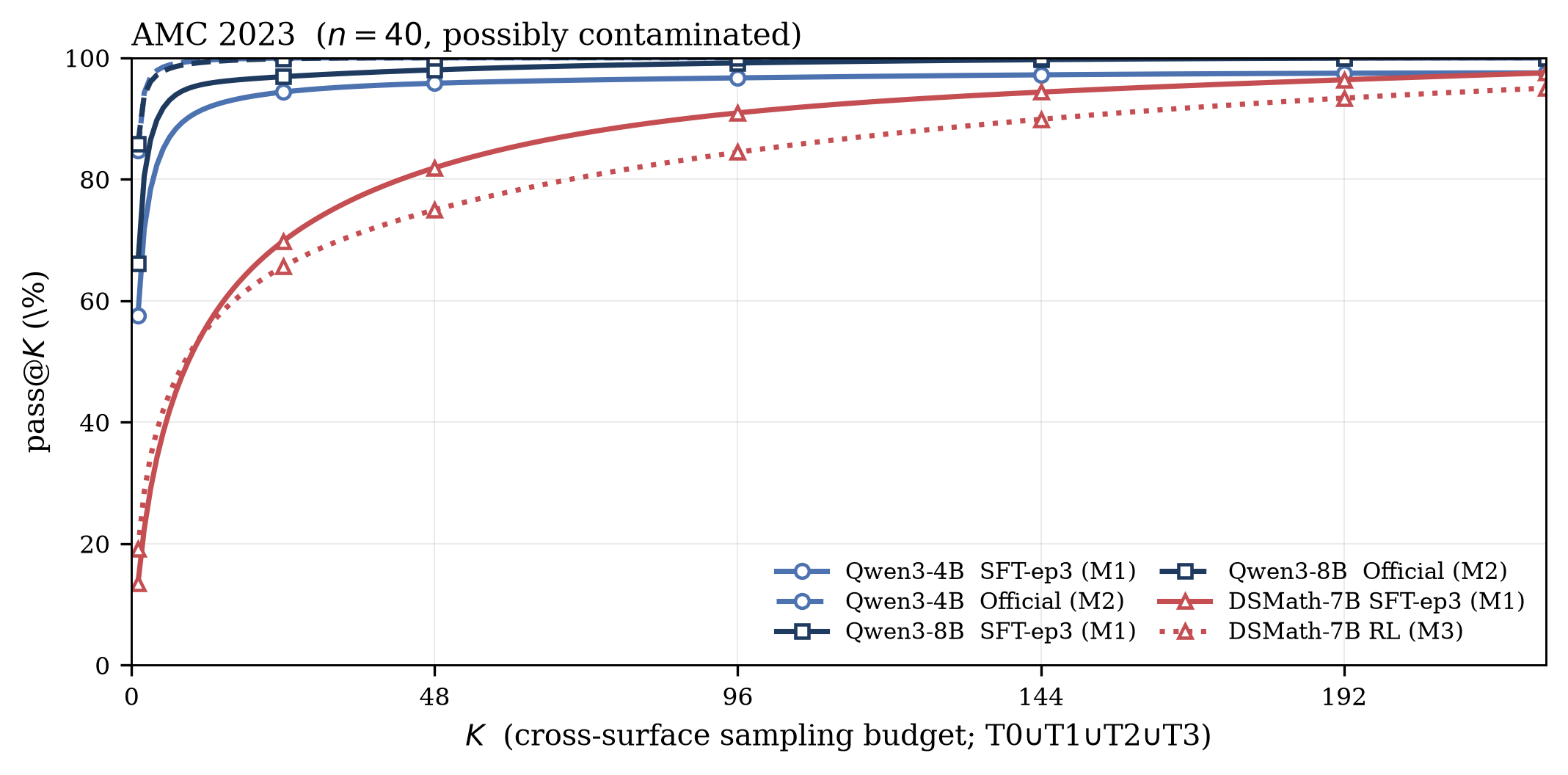}
  \caption{AMC 2023 ($n{=}40$, possibly contaminated)}
  \label{fig:passK_combined_amc}
\end{subfigure}
\caption{\textbf{Cross-mechanism pass@$K$ pooled across
T0+T1+T2+T3} ($K_{\max}{=}224$)}
\label{fig:passK_combined}
\end{figure*}

This appendix reports the full cross-mechanism capability summary
(Table~\ref{tab:capability}) and per-subset numeric anchor points
referenced from \S\ref{sec:mech:ceiling}.
Table~\ref{tab:capability} reports $pass@K$ at
$K\in\{1, 16, 48, 96, 144, 224\}$ for all $15$ checkpoints. The
four subset tables that follow give the values at
$K\in\{1, 48, 96, 144, 224\}$ on the harder AIME 2025+2026 subset
($n{=}60$) and the potentially-contaminated AMC 2023 subset
($n{=}40$) for both Qwen3 families
(Tables~\ref{tab:qwen3_aime},~\ref{tab:qwen3_amc}) and the
DeepSeek-Math-7B family
(Tables~\ref{tab:dsmath_aime},~\ref{tab:dsmath_amc}).
Figure~\ref{fig:passK_combined} provides a combined cross-mechanism
view on AIME (Fig.~\ref{fig:passK_combined_aime}) and AMC
(Fig.~\ref{fig:passK_combined_amc}). It overlays six curves on a
single axis: M1 (off-policy distill, SFT-ep3) for all three families,
M2 (on-policy distill) for the two Qwen3 families, and M3 (GRPO RL)
for DeepSeek-Math-7B. No family has all three mechanisms: M2 exists
only as a released Qwen3 endpoint and M3 only as a released
DeepSeek-Math endpoint.

\begin{table*}[!tb]
\centering
\small
\setlength{\tabcolsep}{5pt}
\begin{tabular}{@{}lllrrrrrrr@{}}
\toprule
\textbf{Family} & \textbf{Stage} & \textbf{Post-training} &
\textbf{pass@1} & \textbf{pass@16} & \textbf{pass@48} & \textbf{pass@96} &
\textbf{pass@144} & \textbf{pass@224} & \textbf{ratio} \\
\multicolumn{3}{l}{\emph{probe set}} & T0 & T0 & T0 & {\footnotesize T1$\cup$T2} & {\footnotesize T0$\cup$T1$\cup$T2} & {\footnotesize $\cup$T3} & \\
\midrule
\multirow{5}{*}{Qwen3-4B}
  & Base    & --                       & 10.5 & 39 & 49 & 62 & 66 & 68 &  6.48$\times$ \\
  & SFT-ep1 & Off-policy distill       & 36.6 & 64 & 70 & 78 & 81 & 85 &  2.32$\times$ \\
  & SFT-ep2 & Off-policy distill       & 38.9 & 66 & 75 & 80 & 81 & 84 &  2.16$\times$ \\
  & SFT-ep3 & Off-policy distill       & 38.7 & 67 & 76 & 82 & 84 & 85 &  2.19$\times$ \\
  & Endpoint & \textbf{Off + On distill} & \textbf{68.4} & \textbf{87} & \textbf{90} & \textbf{93} & \textbf{93} & \textbf{94} & \textbf{1.37$\times$} \\
\midrule
\multirow{5}{*}{Qwen3-8B}
  & Base    & --                       & 12.8 & 44 & 50 & 66 & 68 & 72 &  5.62$\times$ \\
  & SFT-ep1 & Off-policy distill       & 44.6 & 73 & 79 & 84 & 84 & 84 &  1.88$\times$ \\
  & SFT-ep2 & Off-policy distill       & 46.5 & 75 & 80 & 86 & 86 & 88 &  1.89$\times$ \\
  & SFT-ep3 & Off-policy distill       & 46.5 & 77 & 82 & 87 & 89 & 90 &  1.94$\times$ \\
  & Endpoint & \textbf{Off + On distill} & \textbf{69.6} & \textbf{89} & \textbf{90} & \textbf{94} & \textbf{94} & \textbf{94} & \textbf{1.35$\times$} \\
\midrule
\multirow{5}{*}{DSMath-7B}
  & Base    & --                       & 1.1  & 12 & 22 & 34 & 40 & 45 & 40.00$\times$ \\
  & SFT-ep1 & Off-policy distill       & 7.1  & 31 & 44 & 52 & 59 & 60 &  8.45$\times$ \\
  & SFT-ep2 & Off-policy distill       & 7.1  & 31 & 45 & 54 & 59 & 60 &  8.42$\times$ \\
  & SFT-ep3 & Off-policy distill       & 7.4  & 35 & \emph{48} & \emph{54} & \emph{58} & \emph{66} &  8.87$\times$ \\
  & Endpoint & \textbf{Off + RL}        & \textbf{9.7} & 29 & \emph{41} & \emph{51} & \emph{56} & \emph{60} & \textbf{6.18$\times$} \\
\bottomrule
\end{tabular}
\caption{\textbf{Capability across three post-training mechanisms.}
Cross-surface $pass@K$ pooled across T0/T1/T2/T3 surfaces (see
\S\ref{sec:mech:overview}). The \emph{ratio} column reports
$pass@224/pass@1$, a per-checkpoint sample-efficiency multiplier
(computed from unrounded values).}
\label{tab:capability}
\end{table*}

\begin{table*}[!tb]
\centering
\begin{minipage}[t]{0.49\textwidth}
\centering
\small
\setlength{\tabcolsep}{4pt}
\resizebox{\linewidth}{!}{%
\begin{tabular}{@{}llrrrrr@{}}
\toprule
\textbf{Family} & \textbf{Stage} & \textbf{pass@1} &
\textbf{pass@48} & \textbf{pass@96} & \textbf{pass@144} & \textbf{pass@224} \\
\midrule
\multicolumn{2}{l}{\emph{probe set}} & T0 & T0 & {\footnotesize T1$\cup$T2} & {\footnotesize T0$\cup$T1$\cup$T2} & {\footnotesize $\cup$T3} \\
\multirow{3}{*}{Qwen3-4B}
  & Base                  &  2.5 & 27 & 37 & 43 & 47 \\
  & SFT-ep3               & 22.7 & 62 & 72 & 75 & 77 \\
  & \textbf{Endpoint}     & \textbf{55.6} & \textbf{83} & \textbf{88} & \textbf{88} & \textbf{90} \\
\midrule
\multirow{3}{*}{Qwen3-8B}
  & Base                  &  2.9 & 27 & 45 & 48 & 55 \\
  & SFT-ep3               & 28.6 & 72 & 80 & 82 & 83 \\
  & \textbf{Endpoint}     & \textbf{57.2} & \textbf{83} & \textbf{90} & \textbf{90} & \textbf{90} \\
\bottomrule
\end{tabular}%
}
\subcaption{\textbf{AIME 2025+2026 ($n{=}60$).}
On the cross-surface ceiling ($pass@224$) the on-policy distillation
endpoint adds $+13$pp on Qwen3-4B ($77{\to}90$) and $+7$pp on
Qwen3-8B ($83{\to}90$). Single-attempt ($pass@1$) gains
$+33$/$+29$pp are sample-efficiency driven.}
\label{tab:qwen3_aime}
\end{minipage}\hfill
\begin{minipage}[t]{0.49\textwidth}
\centering
\small
\setlength{\tabcolsep}{4pt}
\resizebox{\linewidth}{!}{%
\begin{tabular}{@{}llrrrrr@{}}
\toprule
\textbf{Family} & \textbf{Stage} & \textbf{pass@1} &
\textbf{pass@48} & \textbf{pass@96} & \textbf{pass@144} & \textbf{pass@224} \\
\midrule
\multicolumn{2}{l}{\emph{probe set}} & T0 & T0 & {\footnotesize T1$\cup$T2} & {\footnotesize T0$\cup$T1$\cup$T2} & {\footnotesize $\cup$T3} \\
\multirow{3}{*}{Qwen3-4B}
  & Base                  & 22.4 &  82 & 100 & 100 & 100 \\
  & SFT-ep3               & 62.8 &  98 &  98 &  98 &  98 \\
  & \textbf{Endpoint}     & \textbf{87.6} & \textbf{100} & \textbf{100} & \textbf{100} & \textbf{100} \\
\midrule
\multirow{3}{*}{Qwen3-8B}
  & Base                  & 27.7 &  85 &  98 &  98 &  98 \\
  & SFT-ep3               & 73.2 &  98 &  98 & 100 & 100 \\
  & \textbf{Endpoint}     & \textbf{88.2} & \textbf{100} & \textbf{100} & \textbf{100} & \textbf{100} \\
\bottomrule
\end{tabular}%
}
\subcaption{\textbf{AMC 2023 ($n{=}40$, possibly contaminated).}
AMC predates every model release. The $pass@K{\geq}48$ ceiling is
already at $98$--$100$ after off-policy distillation; the
on-policy lift concentrates at $pass@1$ ($+25$pp 4B, $+15$pp 8B).}
\label{tab:qwen3_amc}
\end{minipage}
\caption{Qwen3 difficulty-stratified capability.}
\end{table*}

\begin{table*}[!tb]
\centering
\begin{minipage}[t]{0.49\textwidth}
\centering
\small
\setlength{\tabcolsep}{5pt}
\resizebox{\linewidth}{!}{%
\begin{tabular}{@{}lrrrrr@{}}
\toprule
\textbf{Stage} &
\textbf{pass@1} & \textbf{pass@48} & \textbf{pass@96} &
\textbf{pass@144} & \textbf{pass@224} \\
\midrule
\multicolumn{1}{l}{\emph{probe set}} & T0 & T0 & {\footnotesize T1$\cup$T2} & {\footnotesize T0$\cup$T1$\cup$T2} & {\footnotesize $\cup$T3} \\
Base                 &  0.0 &  2 &  7 &  8 & 15 \\
SFT-ep3              &  1.1 & \emph{22} & \emph{33} & \emph{37} & \emph{45} \\
\textbf{Endpoint (RL)} & \textbf{1.0} & \emph{20} & \emph{28} & \emph{35} & \emph{37} \\
\bottomrule
\end{tabular}%
}
\subcaption{\textbf{AIME 2025+2026 ($n{=}60$).}
GRPO RL does not exceed SFT-ep3 at any $K\geq 48$, and the point
estimates favour SFT-ep3 throughout ($\Delta pass@224{=}-8$pp,
$45\!\to\!37$). A paired bootstrap over source problems puts every
interval across zero, including the $0.1$pp gap at $pass@1$. Italics
flag the SFT-ep3 $\geq$ RL ordering.}
\label{tab:dsmath_aime}
\end{minipage}\hfill
\begin{minipage}[t]{0.49\textwidth}
\centering
\small
\setlength{\tabcolsep}{5pt}
\resizebox{\linewidth}{!}{%
\begin{tabular}{@{}lrrrrr@{}}
\toprule
\textbf{Stage} &
\textbf{pass@1} & \textbf{pass@48} & \textbf{pass@96} &
\textbf{pass@144} & \textbf{pass@224} \\
\midrule
\multicolumn{1}{l}{\emph{probe set}} & T0 & T0 & {\footnotesize T1$\cup$T2} & {\footnotesize T0$\cup$T1$\cup$T2} & {\footnotesize $\cup$T3} \\
Base                 &  2.8 & 52 & 75 & 88 & 90 \\
SFT-ep3              & 17.0 & \emph{88} & \emph{85} & \emph{90} & \emph{98} \\
\textbf{Endpoint (RL)} & \textbf{22.8} & \emph{72} & \emph{85} & \emph{88} & \emph{95} \\
\bottomrule
\end{tabular}%
}
\subcaption{\textbf{AMC 2023 ($n{=}40$, possibly contaminated).}
RL gains $+5.8$pp at $pass@1$ over off-policy distillation
(single-shot sharpening); this is the only contrast in the table whose
bootstrap interval excludes zero. Beyond it the ordering reverses---
$-15$pp on the T0 pool at $pass@48$, $-2.5$pp on the full
cross-surface pool at $pass@224$ (one problem out of $40$)---within
intervals that do not.}
\label{tab:dsmath_amc}
\end{minipage}
\caption{DeepSeek-Math-7B difficulty-stratified capability.}
\end{table*}

\begin{figure*}[!tb]
\centering
\begin{subfigure}[t]{0.32\linewidth}
  \centering
  \includegraphics[width=\linewidth]{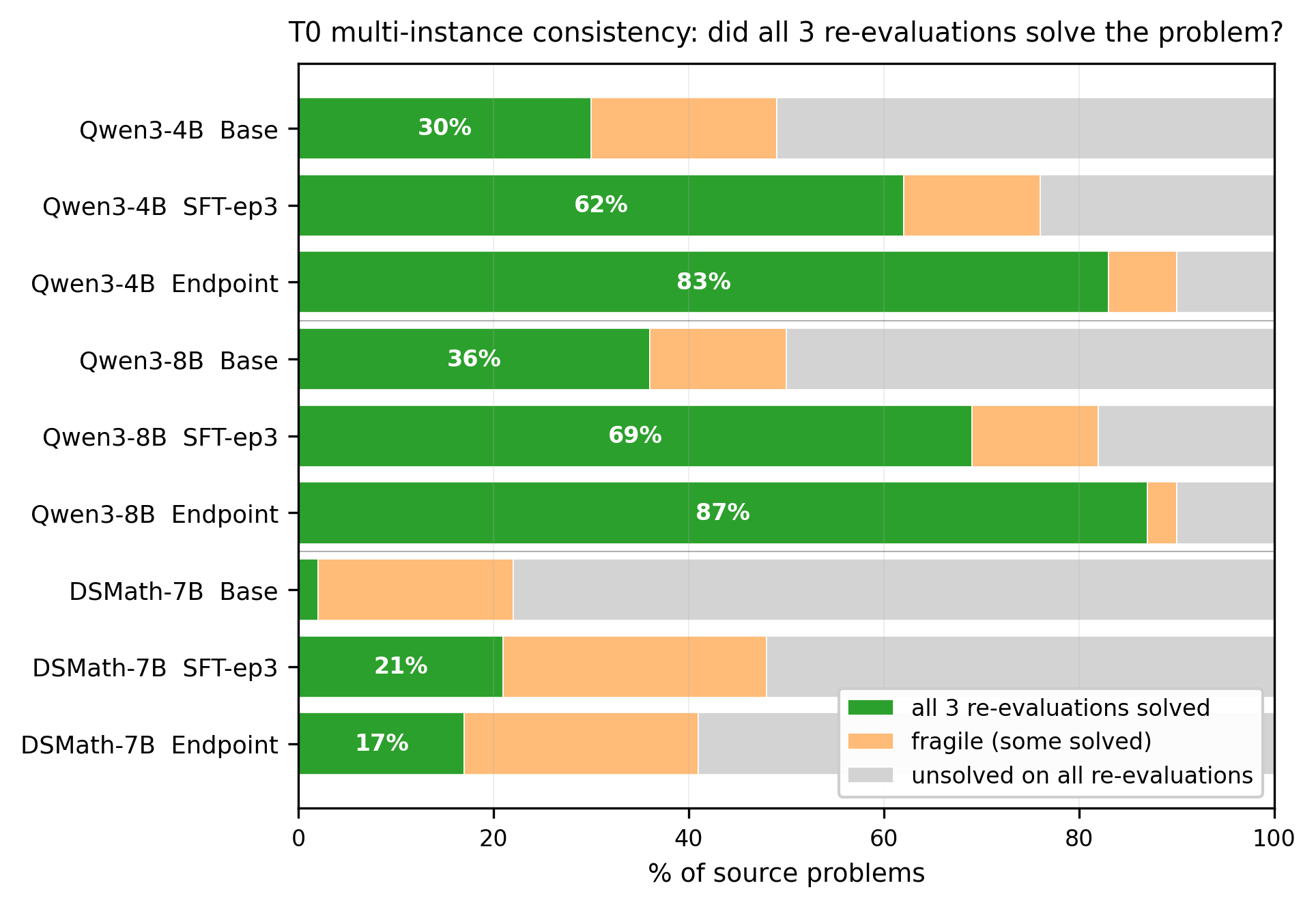}
  \caption{T0: 3 re-evaluations of the original prompt.}
  \label{fig:consistency:t0}
\end{subfigure}\hfill
\begin{subfigure}[t]{0.32\linewidth}
  \centering
  \includegraphics[width=\linewidth]{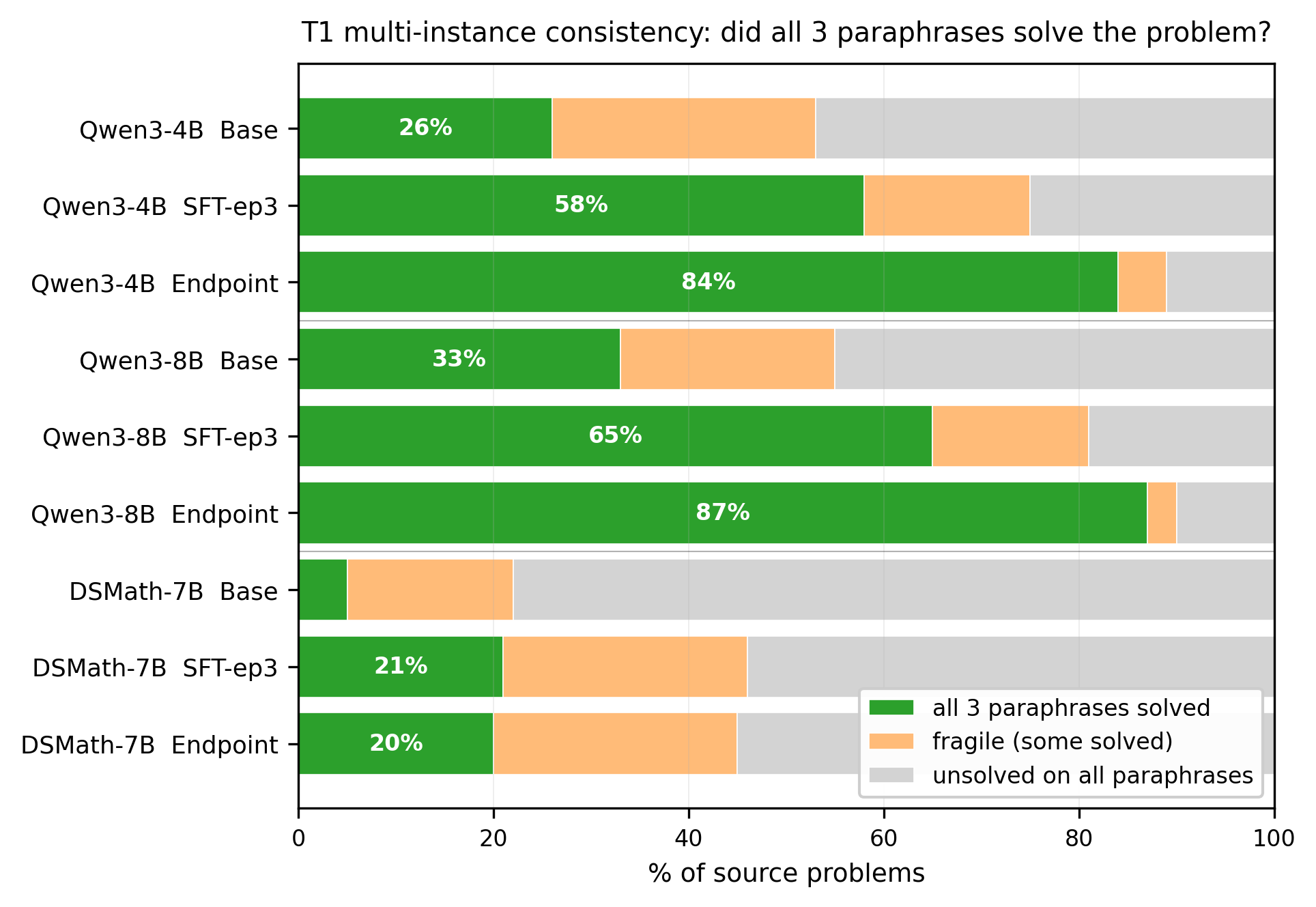}
  \caption{T1: 3 paraphrases.}
  \label{fig:consistency:t1}
\end{subfigure}\hfill
\begin{subfigure}[t]{0.32\linewidth}
  \centering
  \includegraphics[width=\linewidth]{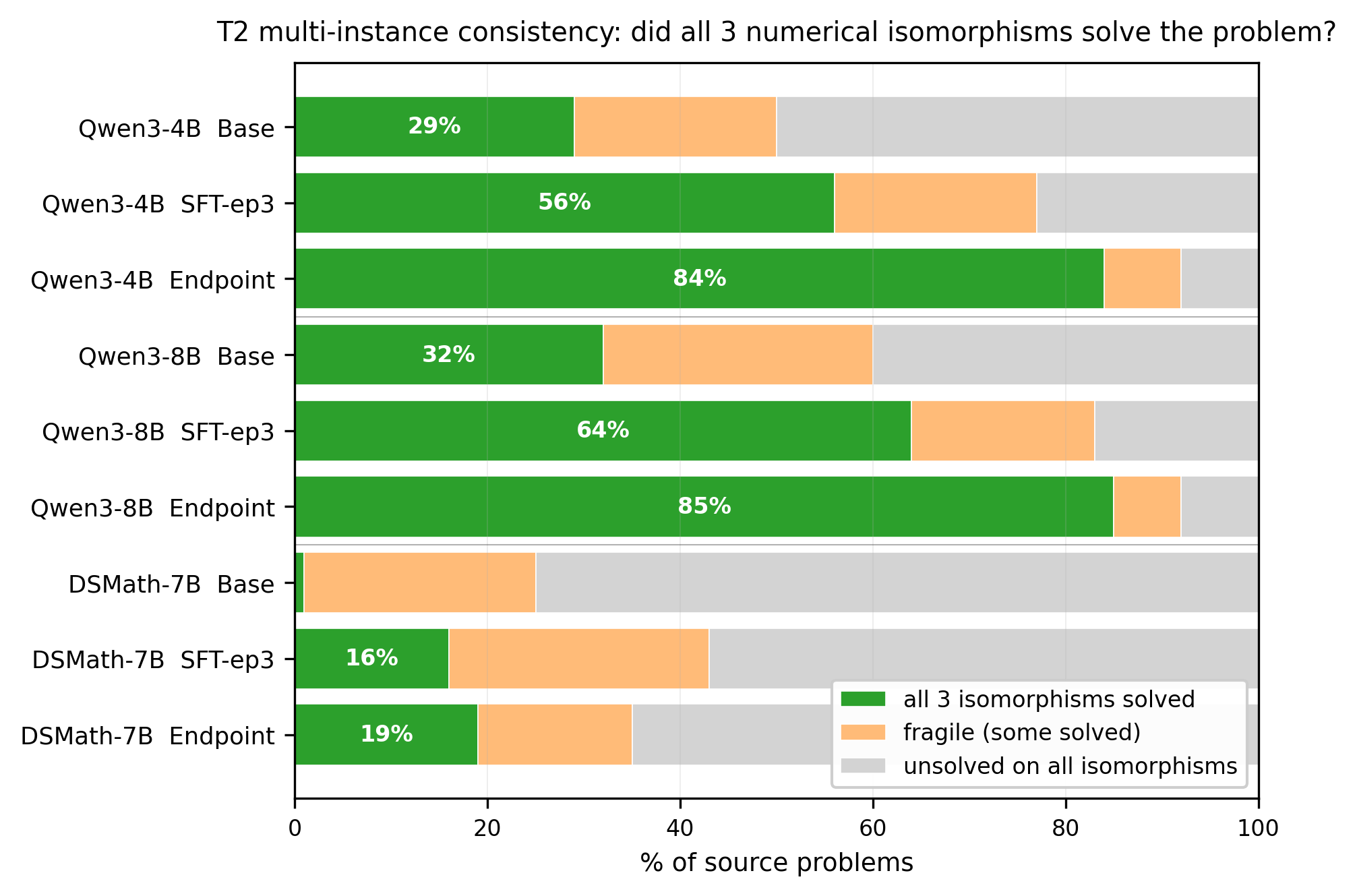}
  \caption{T2: 3 numerical isomorphisms.}
  \label{fig:consistency:t2}
\end{subfigure}
\caption{\textbf{Multi-instance consistency across T0, T1, T2.}
Each panel uses three instances of $K{=}16$ samples per source
problem. T1/T2: three paraphrase/isomorphism variants. T0: the
$K{=}48$ original-prompt pool partitioned into three disjoint
$16$-sample batches so the per-instance budget matches T1/T2.
Green: all three instances solved (at least one correct sample
each); orange: some-but-not-all; grey: unsolved on all three. All
three panels show the same monotone Base$\to$SFT-ep3 consolidation,
with the Qwen3 endpoints further lifting all three transforms; the
DSMath RL endpoint stays within $\pm 4$pp of SFT-ep3 on all three
(T0 $-4$, T1 $-1$, T2 $+3$pp), consistent with
\S\ref{sec:mech:consistency}.}
\label{fig:consistency}
\end{figure*}

\section{Cross-lingual evaluation}
\label{app:multilingual_figs}

\begin{figure*}[!htb]
\centering
\includegraphics[width=\textwidth]{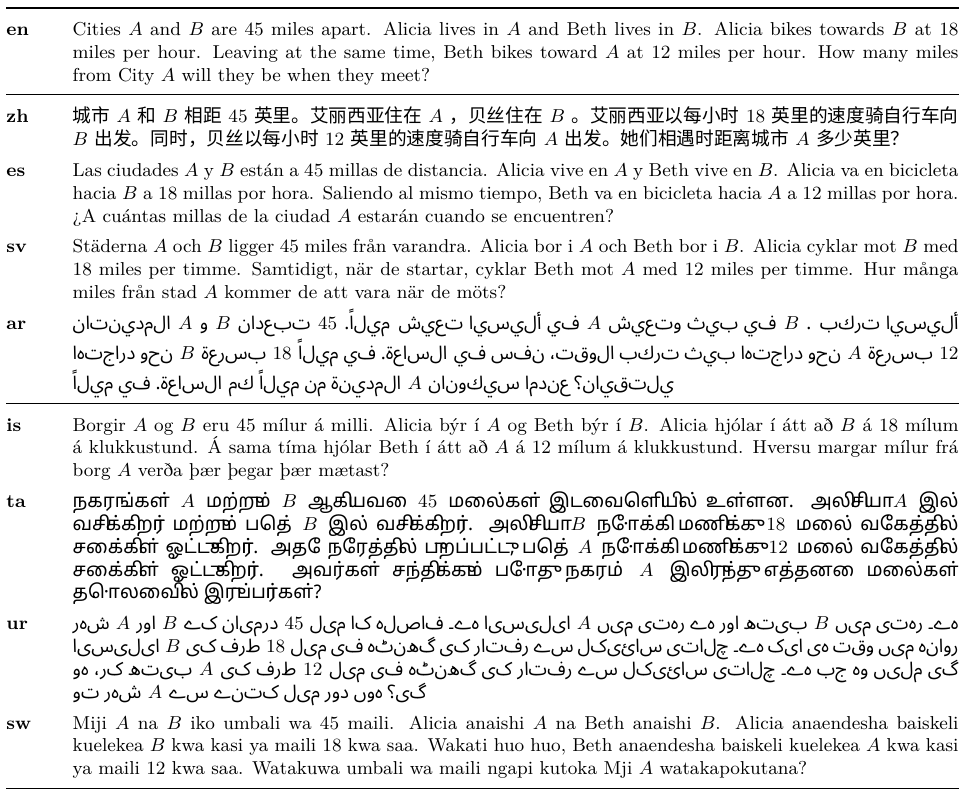}
\caption{\textbf{Representative T3 case.}}
\label{fig:translation_example}
\end{figure*}

\begin{figure*}[!htb]
\centering
\small
\renewcommand{\arraystretch}{1.2}
\begin{tabular}{@{}p{0.97\linewidth}@{}}
\toprule
\textbf{Source problem} (\texttt{amc23\_\_000}, English T0; same item as Figure~\ref{fig:translation_example}):

Cities $A$ and $B$ are $45$ miles apart. Alicia lives in $A$ and Beth lives in $B$. Alicia bikes towards $B$ at $18$ miles per hour. Leaving at the same time, Beth bikes toward $A$ at $12$ miles per hour. How many miles from City $A$ will they be when they meet?

\textbf{Ground truth:} $27$.

\textbf{Model:} Qwen3-4B-SFT-Math-45k-ep3 (off-policy distilled, our SFT-ep3 checkpoint). Sampled at $T{=}0.7$, $\text{top-}p{=}0.95$. Sample $40/48$, $406$ new tokens, scored \texttt{correct}.\\
\midrule
\textbf{Full response (verbatim, $1{,}337$ characters; line breaks preserved):}

{\footnotesize\ttfamily <think>}\\
{\footnotesize Okay, let's see. So, Alicia and Beth are biking towards each other from cities $A$ and $B$, which are $45$ miles apart. Alicia starts from $A$ at $18$ mph, and Beth starts from $B$ at $12$ mph. We need to find out how many miles from $A$ they will meet each other.}

{\footnotesize Hmm, this seems like a relative speed problem. When two objects move towards each other, their speeds add up. So, the combined speed at which they are approaching each other is $18 + 12 = 30$ mph. That makes sense.}

{\footnotesize Now, the total distance between them is $45$ miles. If they're moving towards each other at $30$ mph, how long will it take for them to meet? Well, time is distance divided by speed. So, time $= 45$ miles $/ 30$ mph. Let me calculate that.}

{\footnotesize $45$ divided by $30$ is $1.5$ hours. So, they will meet after $1.5$ hours.}

{\footnotesize Now, we need to find out how far Alicia has traveled in that time. Since speed is distance divided by time, distance $=$ speed $\times$ time. Alicia's speed is $18$ mph, and time is $1.5$ hours. Let me compute that.}

{\footnotesize $18 \times 1.5$. Hmm, $18$ times $1$ is $18$, and $18$ times $0.5$ is $9$. So, $18 + 9 = 27$ miles. So, Alicia will have biked $27$ miles from $A$ when they meet.}

{\footnotesize Alternatively, we could check Beth's distance. Beth's speed is $12$ mph. $12 \times 1.5$ is $18$ miles. $27 + 18 = 45$, which checks out. So that's correct.}

{\footnotesize Therefore, the answer should be $27$ miles from City $A$.}\\
{\footnotesize\ttfamily </think>}

{\footnotesize Answer: $27$}\\
\bottomrule
\end{tabular}
\caption{\textbf{Representative model response with reasoning.}
Off-policy distilled checkpoint (Qwen3-4B SFT-ep3) on the same
source problem as Figure~\ref{fig:translation_example}.}
\label{fig:qa_example}
\end{figure*}

\begin{figure*}[!tb]
\centering
\includegraphics[width=\linewidth]{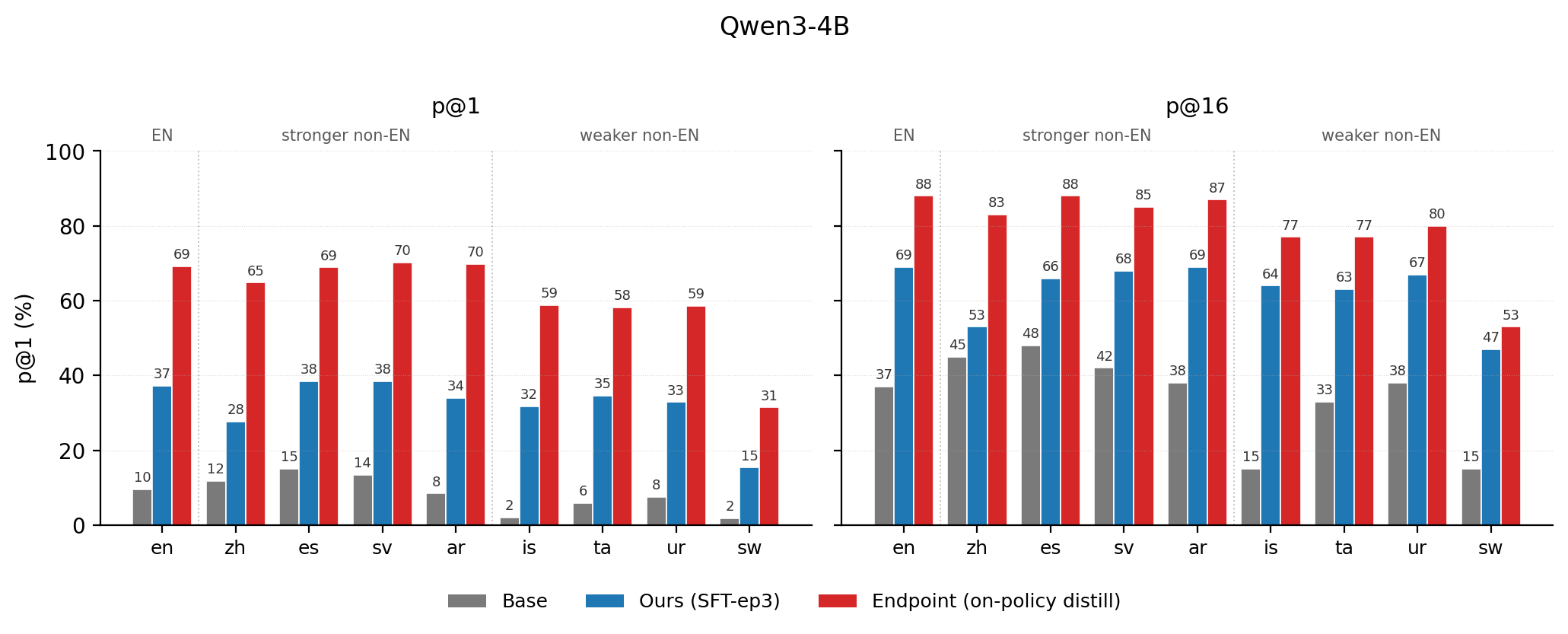}
\caption{\textbf{Qwen3-4B family.} $pass@1$ (left) and $pass@16$
(right) across $9$ languages. Bars: Base / SFT-ep3 (ours) /
Endpoint (on-policy distillation).}
\label{fig:multilingual_qwen3_4b}
\end{figure*}

\begin{figure*}[!tb]
\centering
\includegraphics[width=\linewidth]{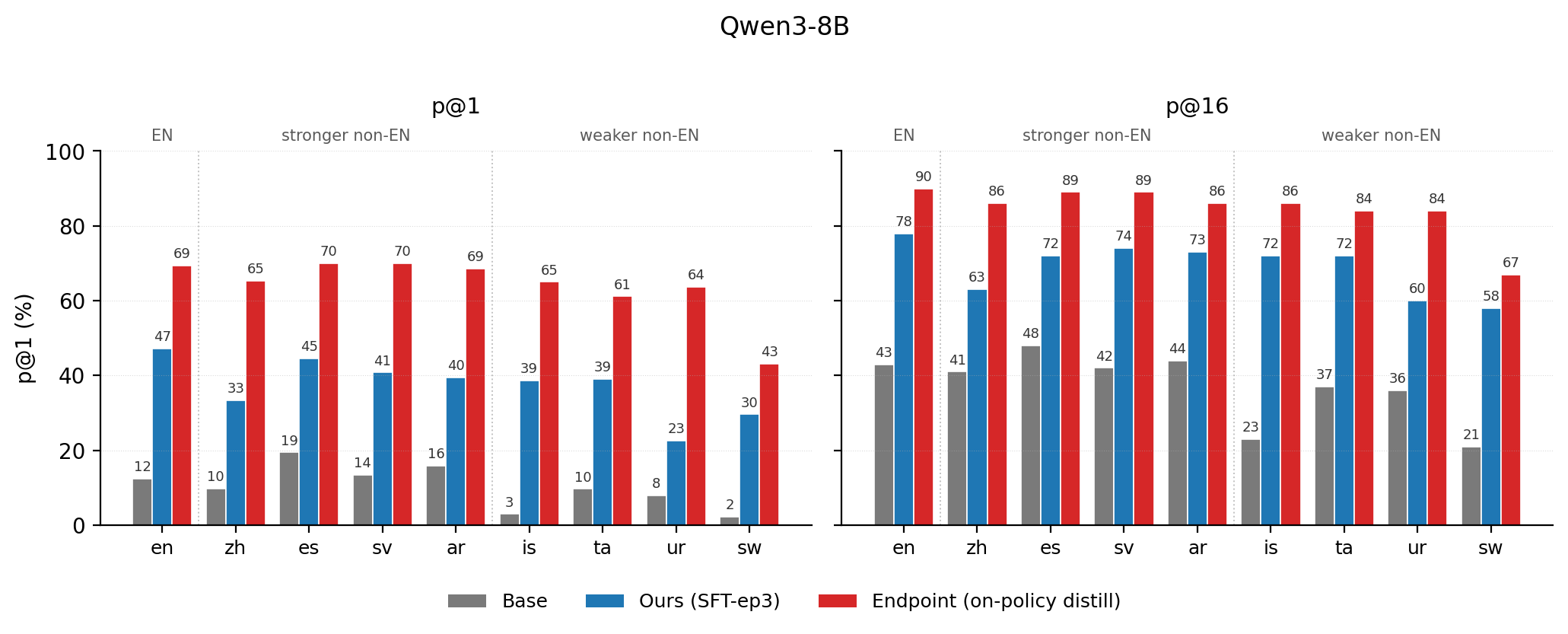}
\caption{\textbf{Qwen3-8B family.} Same layout as
Figure~\ref{fig:multilingual_qwen3_4b}; Base / SFT-ep3 / Endpoint
comparison at 8B scale.}
\label{fig:multilingual_qwen3_8b}
\end{figure*}

\begin{figure*}[!tb]
\centering
\includegraphics[width=\linewidth]{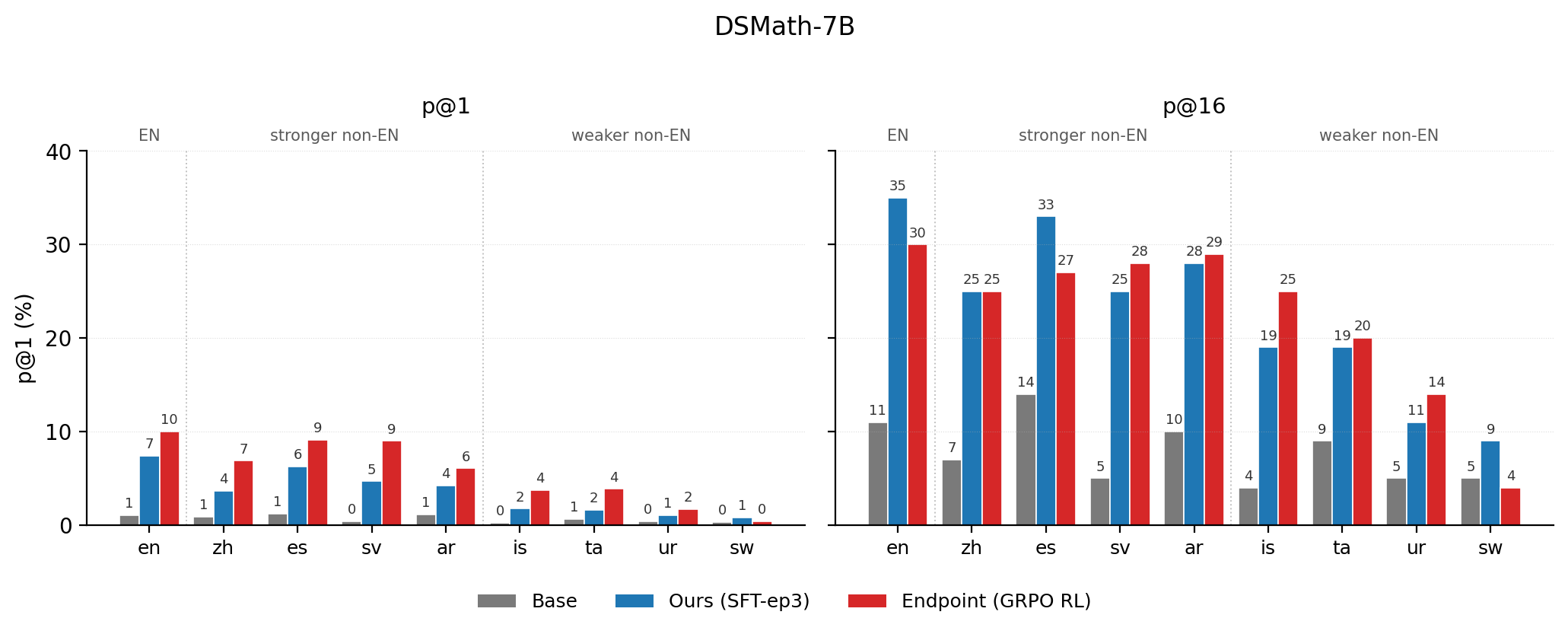}
\caption{\textbf{DeepSeek-Math-7B family.} Endpoint is the
released GRPO RL checkpoint. $y$-axis truncated to $[0, 40]\%$
to keep sub-percent differences visible at smaller $pass@1$
baselines.}
\label{fig:multilingual_dsmath_7b}
\end{figure*}

The aggregated multilingual numbers referenced in
\S\ref{sec:multilingual} are shown in
Table~\ref{tab:multilingual}; per-family breakdowns are in
Figure~\ref{fig:multilingual_qwen3_4b} (Qwen3-4B),
Figure~\ref{fig:multilingual_qwen3_8b} (Qwen3-8B), and
Figure~\ref{fig:multilingual_dsmath_7b} (DeepSeek-Math-7B).

\begin{table*}[!tb]
\centering
\scriptsize
\setlength{\tabcolsep}{4.5pt}
\begin{tabular}{@{}llrrrrrrrrr@{}}
\toprule
\textbf{Model} & \textbf{Metric} &
\multicolumn{1}{c}{\textit{en}} &
\multicolumn{4}{c}{\textit{stronger non-en}} &
\multicolumn{4}{c}{\textit{weaker non-en}} \\
\cmidrule(lr){3-3} \cmidrule(lr){4-7} \cmidrule(lr){8-11}
& & en & zh & es & sv & ar & is & ta & ur & sw \\
\midrule
  \multirow{2}{*}{Qwen3-4B\,Base} & $pass@1$ & 9.6 & 11.9 & 15.0 & 13.5 & 8.4 & 2.1 & 5.9 & 7.6 & 1.8 \\
   & $pass@16$ & 37 & 45 & 48 & 42 & 38 & 15 & 33 & 38 & 15 \\
  \multirow{2}{*}{Qwen3-4B\,SFT-ep3} & $pass@1$ & 37.3 & 27.8 & 38.4 & 38.4 & 34.0 & 31.7 & 34.6 & 32.9 & 15.4 \\
   & $pass@16$ & 69 & 53 & 66 & 68 & 69 & 64 & 63 & 67 & 47 \\
  \multirow{2}{*}{Qwen3-4B\,Endpoint} & $pass@1$ & 69.2 & 64.8 & 68.9 & 70.1 & 69.8 & 58.8 & 58.2 & 58.6 & 31.4 \\
   & $pass@16$ & 88 & 83 & 88 & 85 & 87 & 77 & 77 & 80 & 53 \\
\midrule
  \multirow{2}{*}{Qwen3-8B\,Base} & $pass@1$ & 12.4 & 9.8 & 19.4 & 13.5 & 15.9 & 2.9 & 9.7 & 7.9 & 2.3 \\
   & $pass@16$ & 43 & 41 & 48 & 42 & 44 & 23 & 37 & 36 & 21 \\
  \multirow{2}{*}{Qwen3-8B\,SFT-ep3} & $pass@1$ & 47.2 & 33.3 & 44.6 & 40.8 & 39.5 & 38.6 & 39.0 & 22.6 & 29.6 \\
   & $pass@16$ & 78 & 63 & 72 & 74 & 73 & 72 & 72 & 60 & 58 \\
  \multirow{2}{*}{Qwen3-8B\,Endpoint} & $pass@1$ & 69.3 & 65.3 & 70.1 & 69.9 & 68.6 & 65.0 & 61.1 & 63.7 & 43.2 \\
   & $pass@16$ & 90 & 86 & 89 & 89 & 86 & 86 & 84 & 84 & 67 \\
\midrule
  \multirow{2}{*}{DSMath-7B\,Base} & $pass@1$ & 1.1 & 0.9 & 1.2 & 0.4 & 1.1 & 0.2 & 0.6 & 0.4 & 0.3 \\
   & $pass@16$ & 11 & 7 & 14 & 5 & 10 & 4 & 9 & 5 & 5 \\
  \multirow{2}{*}{DSMath-7B\,SFT-ep3} & $pass@1$ & 7.4 & 3.7 & 6.2 & 4.7 & 4.2 & 1.8 & 1.6 & 1.1 & 0.8 \\
   & $pass@16$ & 35 & 25 & 33 & 25 & 28 & 19 & 19 & 11 & 9 \\
  \multirow{2}{*}{DSMath-7B\,Endpoint} & $pass@1$ & 10.0 & 6.9 & 9.1 & 9.0 & 6.1 & 3.8 & 3.9 & 1.7 & 0.4 \\
   & $pass@16$ & 30 & 25 & 27 & 28 & 29 & 25 & 20 & 14 & 4 \\
\bottomrule
\end{tabular}
\caption{\textbf{Multilingual cross-surface reasoning, $9$ models
$\times\,9$ languages.} Each model contributes two rows: $pass@1$ and $pass@16$.}
\label{tab:multilingual}
\end{table*}

\begin{figure*}[!tb]
\centering
\includegraphics[width=\linewidth]{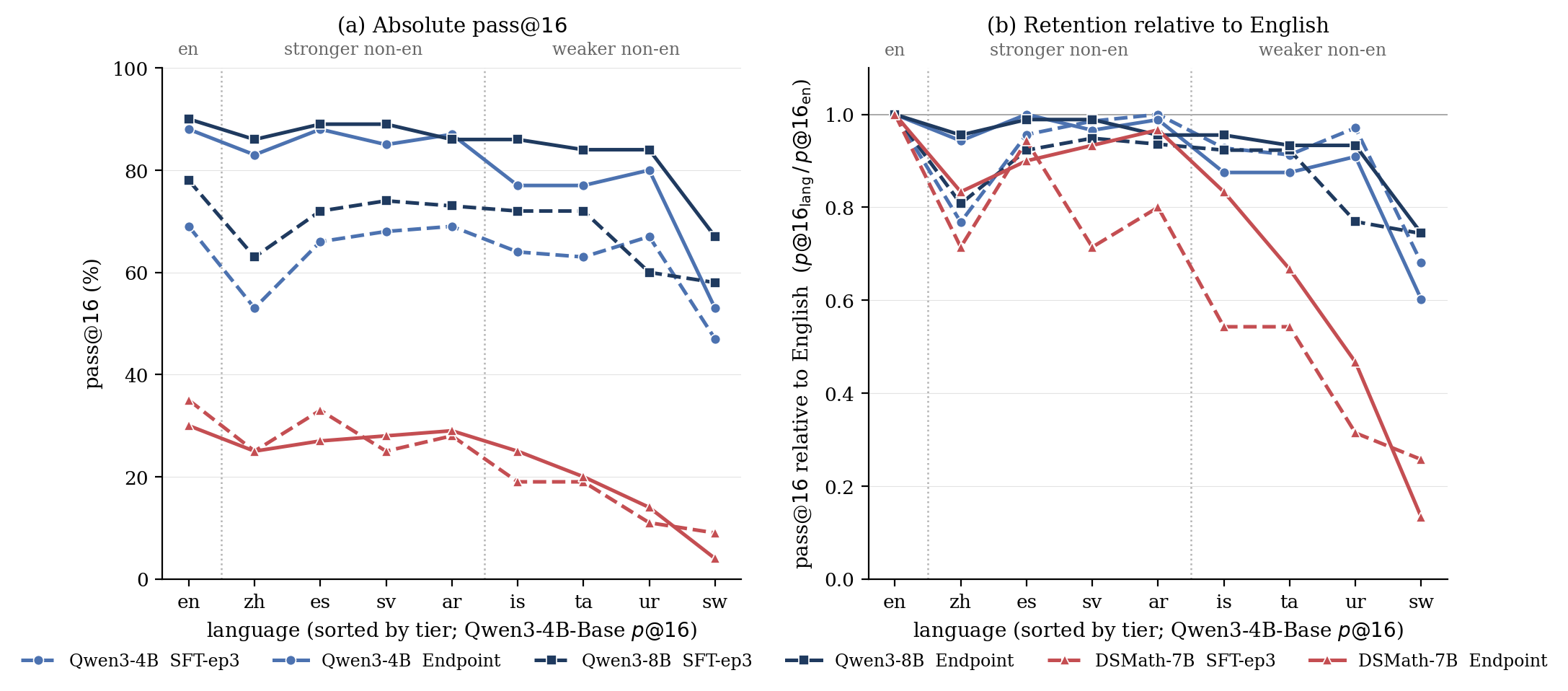}
\caption{\textbf{Multilingual imbalance across mechanisms.}
We plot $pass@16$ over English, stronger non-English languages
({zh, es, sv, ar}), and weaker non-English languages
({is, ta, ur, sw}) for each post-trained model. \textbf{(a)}
Absolute scores show the family-level performance gap. \textbf{(b)}
English-relative retention ($pass@16_{\text{lang}}/pass@16_{\text{en}}$)
normalises across families. Weaker-language gaps persist after
post-training.}
\label{fig:multilingual_imbalance}
\end{figure*}

\begin{figure*}[!tb]
\centering
\begin{subfigure}[t]{0.49\linewidth}
  \centering
  \includegraphics[width=\linewidth]{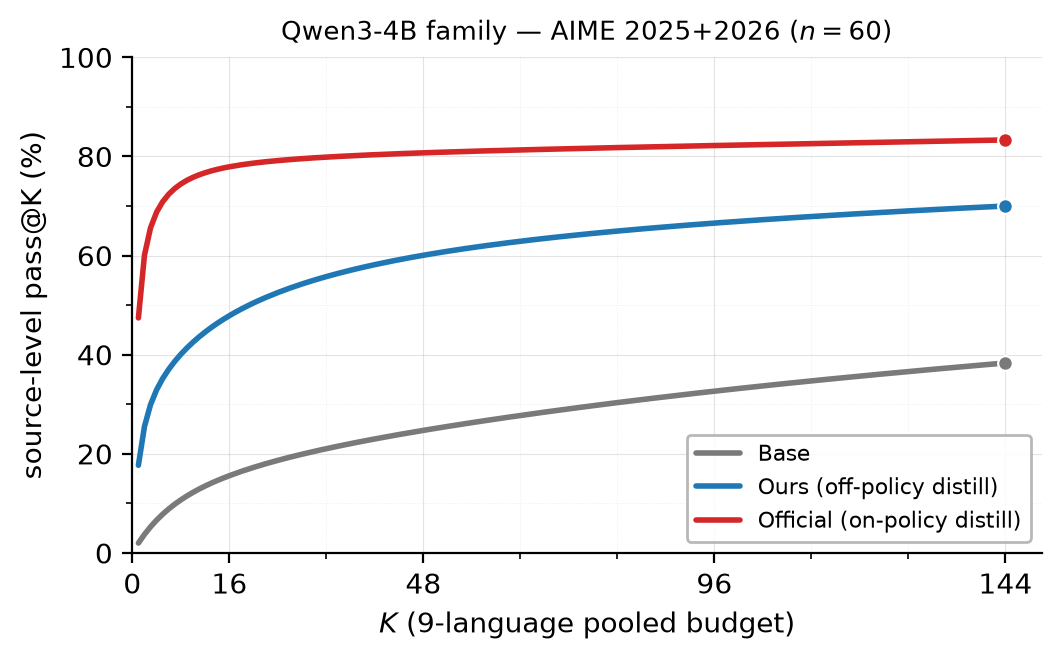}
  \caption{Qwen3-4B, AIME ($n{=}60$).}
  \label{fig:passK_9lang:aime_4b}
\end{subfigure}\hfill
\begin{subfigure}[t]{0.49\linewidth}
  \centering
  \includegraphics[width=\linewidth]{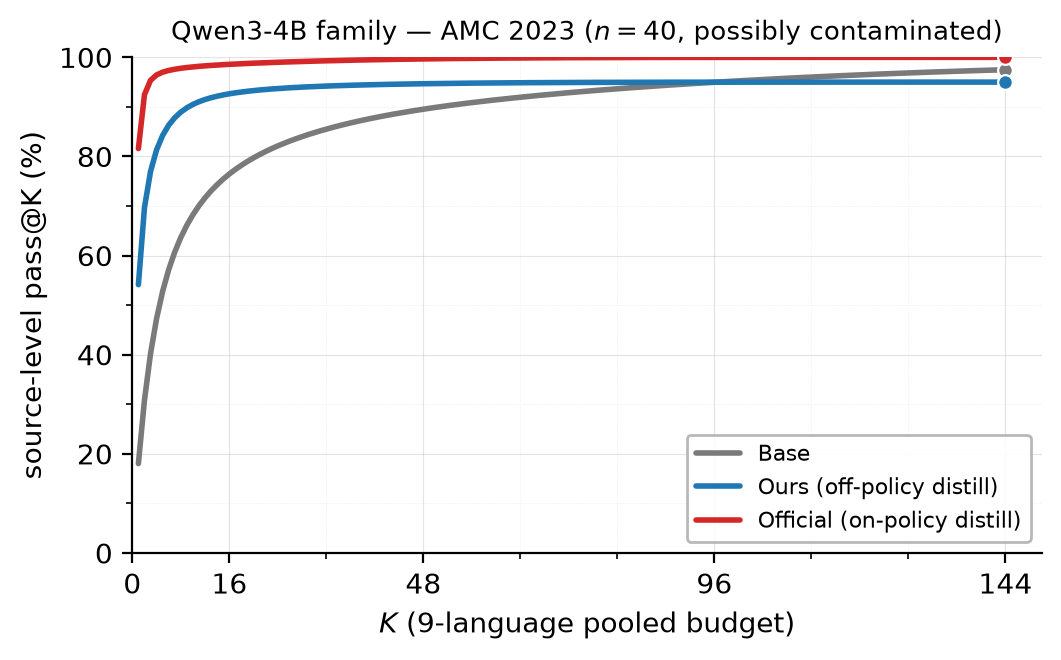}
  \caption{Qwen3-4B, AMC ($n{=}40$).}
  \label{fig:passK_9lang:amc_4b}
\end{subfigure}
\\[2pt]
\begin{subfigure}[t]{0.49\linewidth}
  \centering
  \includegraphics[width=\linewidth]{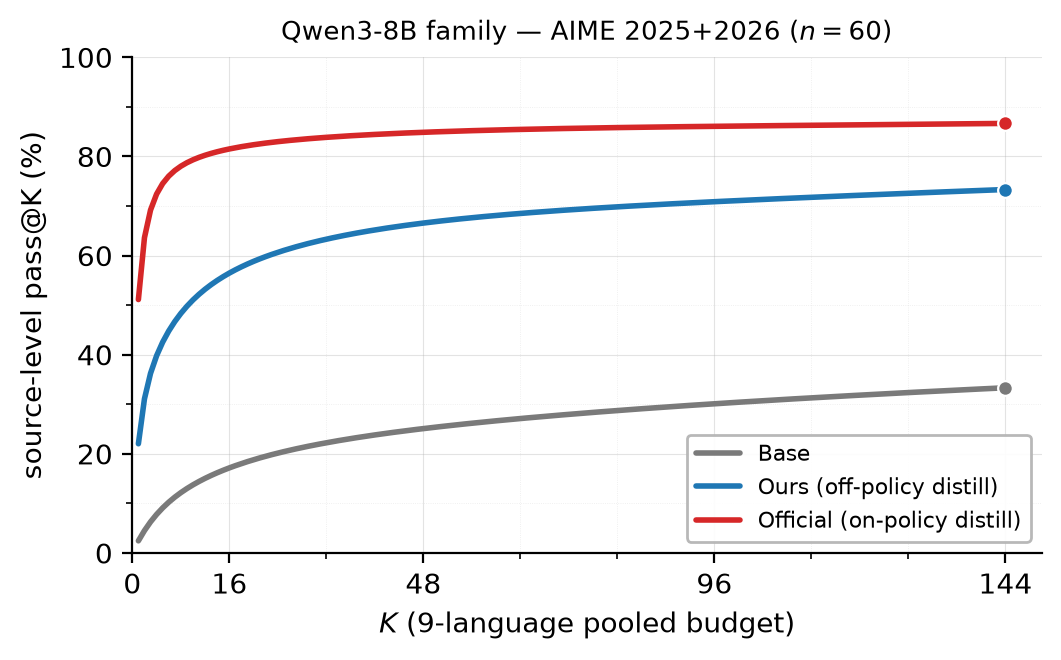}
  \caption{Qwen3-8B, AIME ($n{=}60$).}
  \label{fig:passK_9lang:aime_8b}
\end{subfigure}\hfill
\begin{subfigure}[t]{0.49\linewidth}
  \centering
  \includegraphics[width=\linewidth]{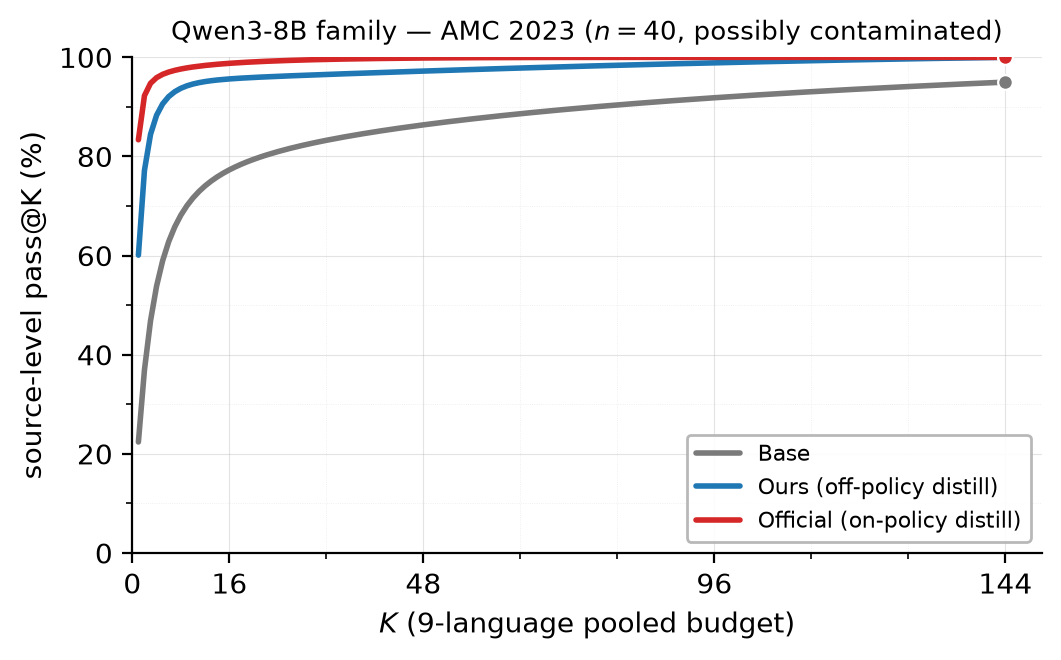}
  \caption{Qwen3-8B, AMC ($n{=}40$).}
  \label{fig:passK_9lang:amc_8b}
\end{subfigure}
\\[2pt]
\begin{subfigure}[t]{0.49\linewidth}
  \centering
  \includegraphics[width=\linewidth]{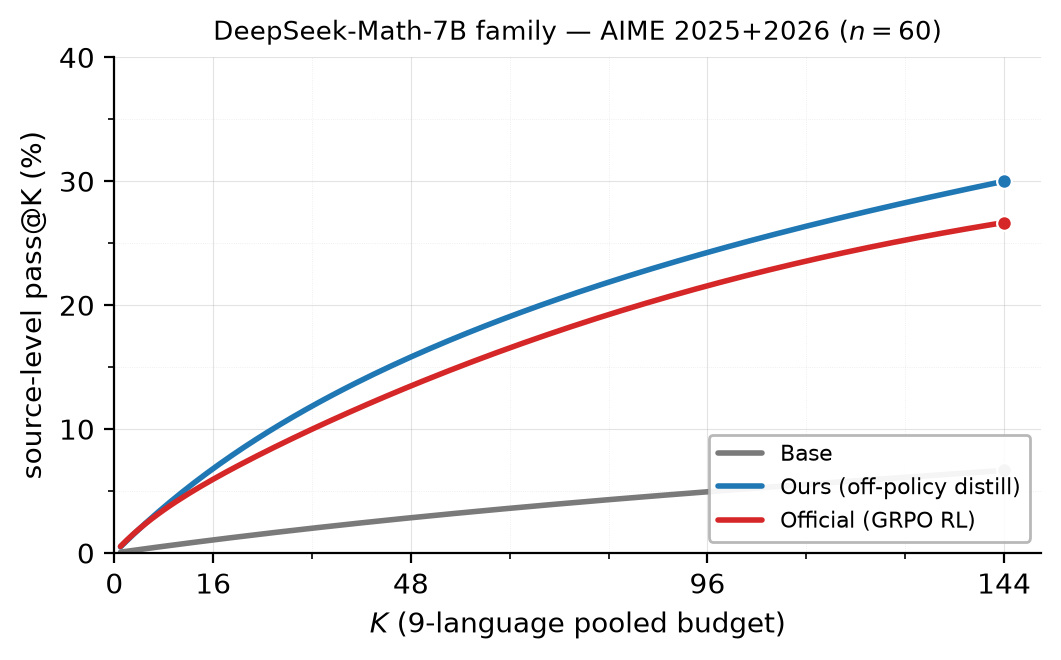}
  \caption{DSMath-7B, AIME ($n{=}60$).}
  \label{fig:passK_9lang:aime_7b}
\end{subfigure}\hfill
\begin{subfigure}[t]{0.49\linewidth}
  \centering
  \includegraphics[width=\linewidth]{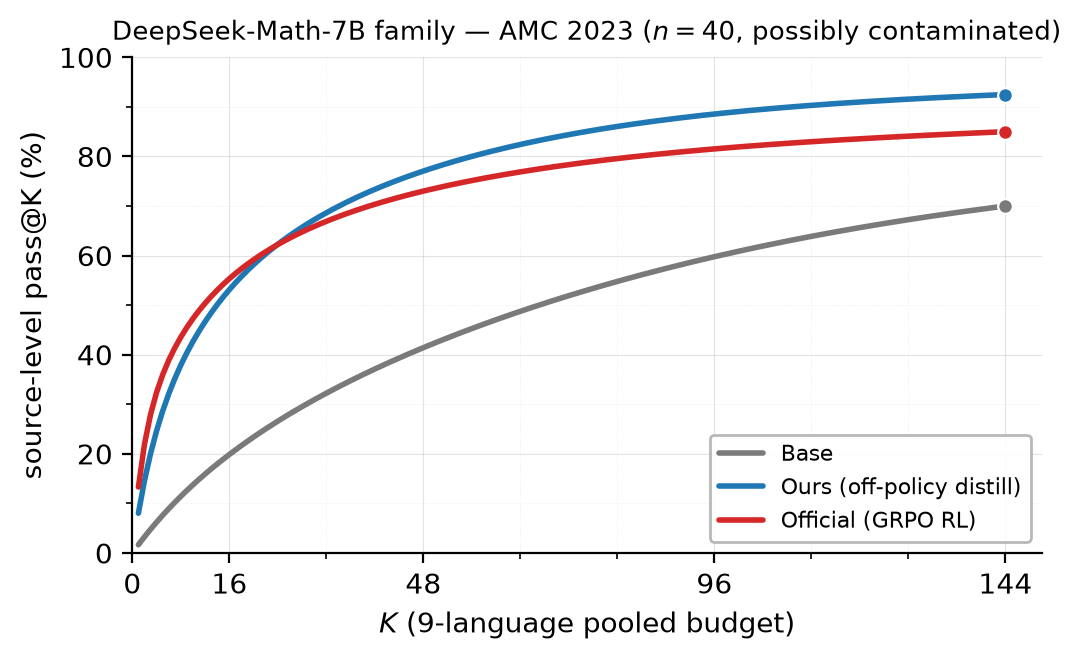}
  \caption{DSMath-7B, AMC ($n{=}40$).}
  \label{fig:passK_9lang:amc_7b}
\end{subfigure}
\caption{\textbf{9-language pooled pass@$K$ split by difficulty
subset.} Each panel pools $9$ languages ($144$ samples per source)
and plots source-level pass@$K$ for $K{=}1\ldots144$. Rows: Qwen3-4B
/ Qwen3-8B / DSMath-7B; columns: AIME 2025+2026 ($n{=}60$) and AMC
2023 ($n{=}40$). On DSMath, GRPO RL leads at small $K$ on AMC
(sharpening) and SFT-ep3 first matches RL at $K{\approx}24$, leading
at every larger $K$. Qwen3 endpoints stay monotone-above
SFT-ep3 throughout.}
\label{fig:passK_9lang}
\end{figure*}

\section{Memorisation: positive controls}
\label{sec:memorisation}
\paragraph{Construction.}
For each base model we sample a $200$-item \emph{seen} set from
its canonical SFT corpus (OpenR1-Math-220k for Qwen3,
dsmath-hybrid for DSMath) and a $200$-item \emph{unseen} held-out
set drawn MD5-disjoint from $220$k. Each base is then fine-tuned
on \emph{only} the seen set for $10$ and $20$ epochs
(full-parameter verl SFT, batch $32$, lr $2{\times}10^{-5}$),
deliberately producing the overfit regime that should maximally
expose memorisation.

\paragraph{Probes.} Each (model, ckpt) cell is evaluated under
three independent probes targeting different memorisation traces:
\textbf{(i) Implicit} --- $40\%$ problem prefix $+$ ``The final
answer is:'', $1024$ tokens, any-of-$16$ boxed-match
\citep{carlini2021extracting};
\textbf{(ii) Direct} --- full problem $+$ ``no reasoning, only
\texttt{\textbackslash boxed\{\}}'', $128$ tokens (answer-recall
under forbidden CoT);
\textbf{(iii) MIA} --- forward-pass Loss / MIN-K\% / Zlib AUC over the
ground-truth answer tokens, conditioned on the problem
\citep{shi2024detecting}; no sampling.
Behavioural cells (i, ii) use one-sided Mann--Whitney on
$n{=}200$ items per split; MIA reports ROC AUC against the
verified split label.

\paragraph{Behavioural probes are noisy.}
Within the same (family, ckpt) cell, qualitative inspection finds
verbatim recitation, length-truncated CoT, and ordinary
problem-solving all co-occurring. Single
$\Delta_{\text{seen}-\text{unseen}}$ values are therefore highly
noisy and inherently unreliable. We exclude behavioural numbers
from the headline range and rely on the deterministic MIA AUC
below.

\paragraph{MIA caveats.}
Forward-pass MIA (Table~\ref{tab:mia}) is one forward pass per
item, so its AUC is stable. Two caveats temper the bounded
range reported in the main text:
(i)~Qwen3 Loss AUC responds to overfit ($+3$--$4$pp from ep$0$
to ep$20$), but DSMath Loss AUC is already $\approx\!0.65$ at
ep$0$ (no overfit) and barely moves with epoch---part of the
DSMath signal therefore reflects baseline corpus statistics
rather than memorisation per se;
(ii)~we did not run a randomised baseline (AUC after relabelling
seen/unseen at random) to anchor what $0.65$ means in this
regime.

\begin{table}
\centering
\small
\setlength{\tabcolsep}{5pt}
\renewcommand{\arraystretch}{1.05}
\begin{tabular}{@{}l c ccc@{}}
\toprule
\textbf{Family} & \textbf{Epochs} &
\textbf{Loss} & \textbf{MIN-K\%} & \textbf{Zlib} \\
& & \multicolumn{3}{c}{\footnotesize \textit{ROC AUC vs.\ unseen}} \\
\cmidrule(lr){3-5}
\multirow{3}{*}{Qwen3-4B}
  & 0  & 0.562 & 0.627 & 0.513 \\
  & 10 & 0.603 & 0.648 & 0.548 \\
  & 20 & 0.596 & 0.644 & 0.540 \\
\midrule
\multirow{3}{*}{Qwen3-8B}
  & 0  & 0.576 & 0.592 & 0.548 \\
  & 10 & 0.613 & 0.606 & 0.583 \\
  & 20 & 0.613 & 0.602 & 0.583 \\
\midrule
\multirow{3}{*}{DSMath-7B}
  & 0  & 0.649 & 0.543 & 0.638 \\
  & 10 & 0.644 & 0.558 & 0.631 \\
  & 20 & 0.645 & 0.555 & 0.633 \\
\bottomrule
\end{tabular}
\caption{\textbf{Forward-pass MIA ROC AUC across the overfit
grid.} $200$ seen vs.\ $200$ unseen items per cell, one forward
pass per item (no sampling). All entries stay within $[.51, .65]$,
below the stronger pretraining-data detection signals reported in
prior work (e.g., Min-K\% Prob reaching AUC $0.88$ for
copyrighted-book detection~\citep{shi2024detecting}).
Qwen3 Loss AUC climbs $+3$--$4$pp from $0$ to $20$ overfit epochs
(an overfit-sensitive signal), while DSMath Loss AUC starts at
$\approx\!0.65$ without overfit and stays essentially flat
(baseline corpus statistics, not memorisation).}
\label{tab:mia}
\end{table}

\end{document}